\documentclass{article}
\usepackage{arxiv}
\usepackage[utf8]{inputenc} 
\usepackage[T1]{fontenc}    
\usepackage{hyperref}       
\usepackage{url}            
\usepackage{booktabs}       
\usepackage{amsfonts}       
\usepackage{nicefrac}       
\usepackage{microtype}      
\usepackage{lipsum}		
\usepackage{graphicx}
\usepackage{natbib}
\usepackage{doi}
\usepackage{multirow}
\usepackage{bbding}
\usepackage{amsmath}
\usepackage{makecell}
\usepackage{tabularx}
\usepackage{subcaption}
\usepackage{booktabs}
\usepackage[dvipsnames]{xcolor}
\usepackage{tcolorbox}

\usepackage{listings} 
\usepackage{tikz}
\usepackage{enumitem}
\usepackage{colortbl}

\newcommand{\dataname}{\textbf{InternVid2}}
\newcommand{\modelname}{\textbf{InternVideo2}}
\newcommand{\annoname}{\textbf{VidCap}}

\title{{\modelname}: Scaling Foundation Models for Multimodal Video Understanding}

\author{Yi Wang$^{*1}$, Kunchang Li$^{*3,1}$, Xinhao Li$^{*2,1}$, Jiashuo Yu$^{*1}$, Yinan He$^{*1}$, Chenting Wang$^{*2,1}$\\
\textbf{Guo Chen$^{2,1}$, Baoqi Pei$^{1}$, Ziang Yan$^{1}$, Rongkun Zheng$^{1}$, Jilan Xu$^{1}$, Zun Wang$^{1}$}\\
\textbf{Yansong Shi$^{1}$, Tianxiang Jiang$^{1}$, Songze Li$^{1}$, Hongjie Zhang$^{1}$, Yifei Huang$^{1}$}\\
\textbf{Yu Qiao$^{\dagger 1}$, Yali Wang$^{\dagger 3,1}$, Limin Wang$^{\dagger 2,1}$}\\\
$^1$OpenGVLab, Shanghai AI Laboratory, Shanghai, China \quad $^2$Nanjing University, Nanjing, China \\$^3$Shenzhen Institutes of Advanced Technology, CAS, Shenzhen, China \\ 
 \\
\small{\url{https://github.com/OpenGVLab/InternVideo/tree/main/InternVideo2}} \\
 \hspace{-0.25cm}
}

\definecolor{darkGreen}{RGB}{92, 148, 110}
\definecolor{myblue}{RGB}{14, 121, 178}
\newcommand{\orange}[1]{\textcolor{orange}{#1}}
\newcommand{\violet}[1]{\textcolor{violet}{#1}}
\newcommand{\darkGreen}[1]{\textcolor{darkGreen}{#1}}
\newcommand{\myblue}[1]{\textcolor{myblue}{#1}}
\newcommand{\gray}[1]{\textcolor{gray}{#1}}

\renewcommand{\shorttitle}{{\modelname}: Scaling Foundation Models for Multimodal Video Understanding}

\renewcommand{\cite}{\citep}

\hypersetup{
pdftitle={{\modelname}: Scaling Foundation Models for Multimodal Video Understanding},
pdfsubject={computer vision, machine learning},
pdfauthor={Shanghai AI Lab},
pdfkeywords={Video representation, Multimodal learning, },
}

\begin{document}
\maketitle

\begin{figure*}[th]
    \centering
    \includegraphics[width=1\textwidth]{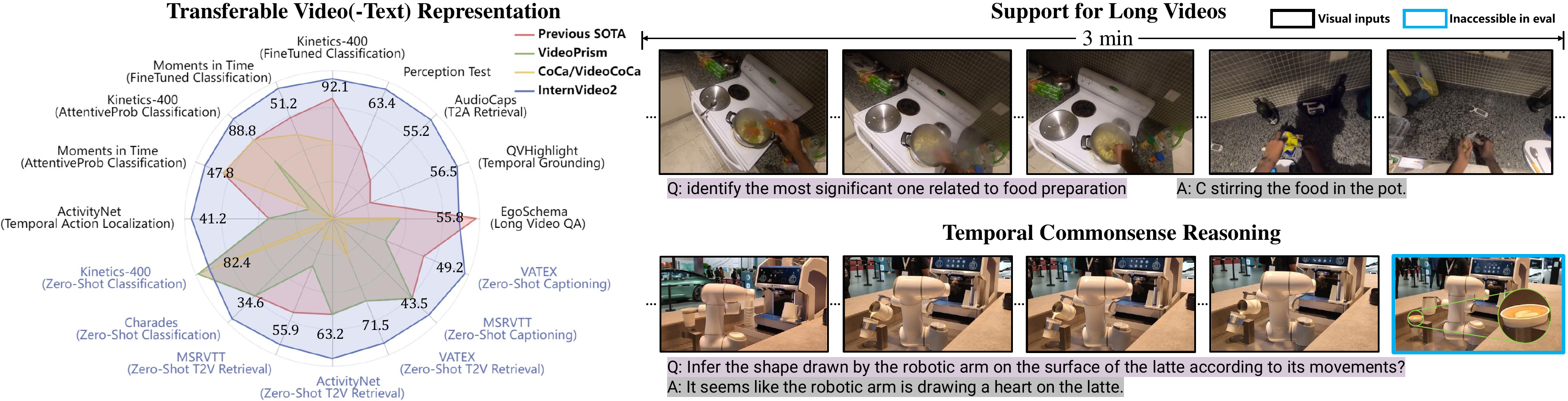}
    \vspace{-0.3cm}
    \caption{{\modelname} yields strong transferable visual and visual-linguistic representations across a total of 70 video understanding tasks, ranging from action recognition, video-text understanding, to video-centric dialogue. It also exhibits capability of long-form video understanding and procedure-aware reasoning.}
    \label{fig:teaser}
\end{figure*}

\begin{abstract}
We introduce {\modelname}, a new family of video foundation models (ViFM) that achieve the state-of-the-art results in video recognition, video-text tasks, and video-centric dialogue. 
Our core design is a progressive training approach that unifies the masked video modeling, crossmodal contrastive learning, and next token prediction, scaling up the video encoder size to 6B parameters. At the data level, we prioritize spatiotemporal consistency by semantically segmenting videos and generating video-audio-speech captions. This improves the alignment between video and text. Through extensive experiments, we validate our designs and demonstrate superior performance on over 60 video and audio tasks. Notably, our model outperforms others on various video-related dialogue and long video understanding benchmarks, highlighting its ability to reason and comprehend longer contexts.  

{
  \renewcommand{\thefootnote}%
    {\fnsymbol{footnote}}
  \footnotetext[0]{*Equal contribution. $\dagger$Corresponding authors.} 
  }
\end{abstract}

\section{Introduction} \label{sec:intro}

Learning transferrable spatiotemporal representations is a critical research area in computer vision, holding diverse applications across domains such as video search \cite{gabeur2020multi}, game control \cite{bruce2024genie}, robotic learning \cite{palme}, self-driving \cite{zablocki2022explainability}, and scientific studies~\cite{team2023gemini}. Recently, the advancement of Large Language Models (LLMs)~\cite{gpt3,gpt4,llama1,llama2} and their multimodal variations (MLLMs) \cite{gpt4v,mmgpt,llava,team2023gemini} have had a profound impact on vision research and other disciplines. Embedding videos effectively into these large models and harnessing their capabilities to enhance video understanding performance has emerged as pivotal tasks~\cite{li2023videochat,videochatgpt}.

Previous research has identified several effective learning schemes for video representations, including reconstructing videos with masked inputs~\cite{he2022masked,videomae,wang2023videomae,st_mae}, aligning videos with languages \cite{cpd,videoclip,videococa,umt}, and predicting the next token using videos \cite{flamingo,sun2023emu1,li2023mvbench}. These approaches have turned out to be complementary and can be unified through a progressive training scheme. Notably, methods such as InternVideo~\cite{wang2022internvideo}, UMT~\cite{umt}, and VideoPrism~\cite{videoprism} have utilized a two-stage training approach involving masked reconstruction and multimodal contrastive learning, leading to improved performance in downstream tasks. Following this line, we aim to further extend this progressive training scheme by incorporating video-based next token prediction and scaling the entire training process, including models and data, to build a new family of video foundation models. 

The proposed video foundation model, coined as {\modelname}, is built through a progressive training scheme. The learning involves three stages: (1) capturing spatiotemporal structure via unmasked reconstruction, (2) aligning with semantics from other modalities, and (3) enhancing its open-ended dialogue power through next token prediction.
In the initial stage, the model learns to reconstruct the unmasked video tokens, allowing the video encoder to develop basic spatiotemporal perception capability. To estimate the existing tokens, vision encoders (InternViT \cite{chen2023internvl} and VideoMAE-g \cite{wang2023videomae}) trained differently are employed as proxies.
In the next stage of crossmodal learning, the architecture is expanded to include audio and text encoders. This not only improves the alignment between videos and text but also endows {\modelname} the ability to handle video-audio tasks. By incorporating these additional modalities, the model's understanding of videos is enriched and aligned with their semantics.
Finally, in the next-token prediction stage, a video-centric dialogue system and the corresponding instruction-finetuning dataset are built to further tune the {\modelname}. By connecting {\modelname} to LLMs, the video encoder is further updated through next-token prediction training, enhancing its ability for open-ended tasks such as VQA and video description. 

For the training of {\modelname}, we emphasize the spatiotemporal consistency and labeling quality in the data. We build a large-scale multimodal video-centric dataset consisting of 402M data entries, which includes 2M videos, 50M video-text pairs (from WebVid~\cite{webvid} and InternVid~\cite{wang2023internvid}), 50M video-audio-speech-text pairs (InternVid2), and 300M image-text pairs.
Specifically, for InternVid2, we segment videos into clips semantically and focus on recalibrating the clip descriptions using three modalities: audio, video, and speech. We first generate captions for these three modalities separately. Then individual captions are fused together to create a more comprehensive description, which will improve the model's ability to comprehend and interpret the video accurately.

We evaluate {\modelname} across a wide range of video-related tasks. These tasks span from basic spatiotemporal perception, such as action recognition, to high-level reasoning tasks, such as long video or procedure-aware question-answering (QA), as given in Fig. \ref{fig:teaser}. The results (in Sec. \ref{sec:exp}) demonstrate that {\modelname} achieves the state-of-the-art performance on multiple tasks and is able to analyze and reason over sequences of actions. This top performance signifies its capability to effectively capture and understand video content. These empirical findings validate that {\modelname} could serves as a general video encoder for future exploration in video understanding. In summary, our contributions to video understanding are as follows.
\begin{itemize}[leftmargin=*]
\item[$\bullet$] This paper introduces {\modelname}, a competitive family of video foundation models that leverages masked reconstruction, crossmodal contrastive learning, and next token prediction to make model perceptive, semantic, and capable of reasoning in video understanding. 
\item[$\bullet$] {\modelname} achieves the state-of-the-art performance for more than 60 video / audio tasks. Our model demonstrates superior performance in video-related dialogue and long video understanding, highlighting its potential in modeling high-level world knowledge.
\item[$\bullet$] We provide an enhanced dataset to train {\modelname}. This includes the validation and incorporation of audio data during training, as well as the improved captioning method. These improvements result in significant enhancements in model performance and generalization ability.
\end{itemize}
\section{Related Work} \label{sec:relate}
\noindent \textbf{Video Foundation Models.}
Studies on learning video foundation models become increasingly crucial considering its wide applications~\cite{cpd,videoclip,umt,wang2022internvideo,videoprism,wang2023masked,videococa,st_mae,wang2023allinone,videomae,wang2023videomae}. Typical methods in building video foundation models (ViFM) include video-text contrastive learning~\cite{cpd,videoclip,wang2022internvideo}, masked video modeling~\cite{videomae,wang2023videomae,wang2022internvideo,violet}, and next token prediction \cite{flamingo,sun2023emu1,sun2023emu2}. Specifically, All-in-one \cite{wang2023allinone} utilized a single backbone with unified multiple pretraining objectives. On the other hand, UMT~\cite{umt} combined masked modeling with video-text contrastive learning, demonstrating strong performance in both action recognition and video-language tasks. Another approach is mPLUG-2~\cite{xu2023mplug}, which introduced a new design for modulating different modalities. It shared a common module across modalities to enhance relations while incorporating modality-specific modules for discrimination.
In addition to video-text pretraining, researchers have also explored the use of audio information in videos to improve performance. MERLOT Reserve \cite{zellers2022merlot} learned video representations using a large-scale dataset of video-speech-transcript pairs. VALOR \cite{chen2023valor} employed independent encoders for video, audio, and text and trains a joint visual-audio-text representation. VAST \cite{chen2024vast} constructed an audio-visual-speech dataset and develops a multimodal backbone that excels in video-audio-related tasks. VideoPrism \cite{videoprism} combined video-text contrastive learning and video token reconstruction on a combination of public and proprietary videos, achieving leading results across various video tasks.

\noindent \textbf{Multimodal Large Language Models.} With advances in large language models (LLMs)~\cite{devlin2018bert,t5,gpt3}, their multimodal versions (MLLMs) is becoming popular as it can handle open-world tasks. Seminal works like Flamingo~\cite{flamingo} showed outstanding zero/few-shots performances over a range of multimodal tasks~\cite{vqa,fickr,msrvtt,okvqa}. Public MLLMs~\cite{minigpt4,llava,mmgpt}
such as LLaVA~\cite{llava} and InstructBLIP~\cite{instructblip} proposed to use visual instruction-tuning data to improve the visual dialogue ability. Some video-centric MLLMs have been proposed, such as VideoChat~\cite{li2023videochat}, VideoChatGPT~\cite{videochatgpt} and Valley~\cite{valley}, by using instruction data to connect video encoders to LLMs for open-world video understanding. 
\begin{figure*}[t]
    \centering
    \includegraphics[width=1\textwidth]{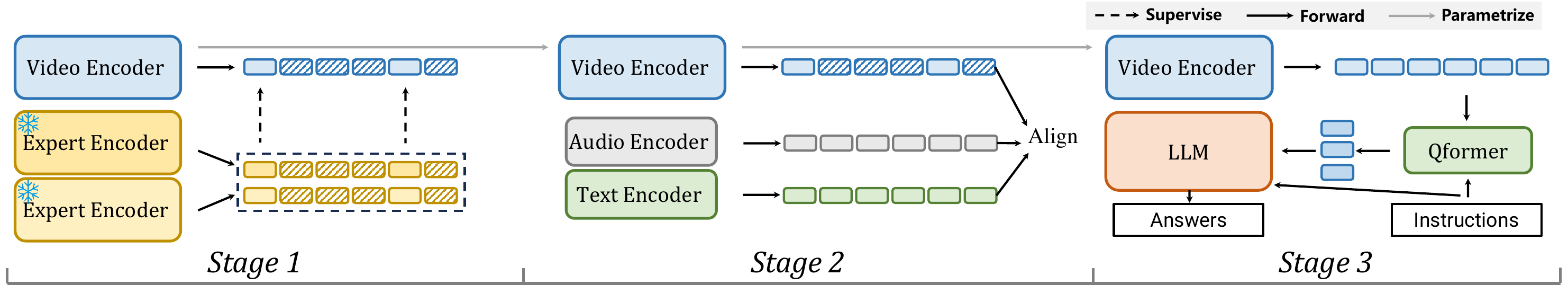}
    \vspace{-0.3cm}
    \caption{Framework of {\modelname}. It consists of three consecutive training phases: unmasked video token reconstruction, multimodal contrastive learning, and next token prediction. In stage 1, the video encoder is trained from scratch, while in stages 2 and 3, it is initialized from the version used in the previous stage.
    }
    \vspace{-3mm}
    \label{fig:frame}
\end{figure*}

\section{Method} \label{sec:method}

We learn {\modelname} in three stages, illustrated in Fig. \ref{fig:frame}. The stages include spatiotemporal token reconstruction, video-audio-speech-language contrastive learning, and connecting to a large language model (LLM) for joint training.

\paragraph{Video Encoder.} The video encoder used in {\modelname} follows the Vision Transformer (ViT) \cite{dosovitskiy2020image} and includes additional projection layers for distillation. Inspired by previous works \cite{chen2023internvl,coca}, we introduce attention pooling to the ViT.
For input videos, we sparsely sample 8 frames~\cite{tsn} and perform a 14$\times$14 ($h \times w$) spatial downsampling. These spatiotemporal tokens are then concatenated with a class token and combined with 3D position embeddings. The details of ViT-6B architecture are given in Supp.

\subsection{Stage1: Reconstructing Unmasked Video Tokens}

We exploit two expert models to guide the video encoder to conduct token-level reconstruction for unmasked areas.
Specifically, we adopt InternVL-6B~\cite{chen2023internvl} and VideoMAEv2-g~\cite{wang2023videomae} to transfer unmasked knowledge via simple projection layers. 
When training, we input the full videos into different teachers and mask out 80\% of the tokens frame by frame, under the semantic guidance of the multimodal model InternVL and motion-aware model VideoMAEv2. We only align the unmasked tokens, by minimizing their mean squared error (MSE) between student and teachers. The learning objective is to reconstruct the remaining tokens as:
\begin{equation}
    \mathcal{L} = \frac{1}{Z}\sum_{}{(\alpha_1|f^V(\mathbf{V}_p)-h(\mathbf{V}_p)|^2+\alpha_2|f^V(\mathbf{V}_p)-g(\mathbf{V}_p)|^2)},
\end{equation}
where $f^V$, $h$, and $g$ are our video encoder, InternViT-6B~\cite{chen2023internvl}, and ViT-g of VideoMAEv2, respectively. $p$ stands for the token index and $f(\mathbf{V}_p)$ is the corresponding token extracted by {\modelname} for input video $\mathbf{V}$. $Z$ is the normalization factor. $\alpha_1$ and $\alpha_2$ balance the influence between the employed models.

In our implementation, we randomly initialize the video encoder and then align its outputs from different layers (transformed by learnable multilayer perceptrons) to those of expert models. Specifically, we align: 1) the last $6$ layers of InternVL, 2) the last $4$ layers of VideoMAEv2, and 3) the final output token of InternVL. These alignments are made to the corresponding outputs of the video encoder using the $l_2$ norm. The different loss terms are simply summed for optimization. After pretraining, we drop those projection layers and only use the basic encoder. Compared with only using the multimodal model in UMT and VideoPrism, our strategy makes the vision encoder multimodal-friendly as well as enhances its temporal sensitivity for action modeling.

\subsection{Stage 2: Aligning Video to Audio-Speech-Text}

We exploit the correspondence between video and audio, speech, and text to encourage {\modelname} to learn more semantics. 
In practice, {\modelname} has a huge video encoder, and its employed audio and text encoders are relatively lightweight. The used audio encoder is a 12-layer transformer initialized with BEATs~\cite{beats} (90M). It takes as input 64-dimensional log Mel filterbank spectrograms, generated using a 25ms Hamming window, from 10-second-long clips (padded with zeros).
For the text and speech encoders, we initialize the text encoder and multimodal decoder using BERT-Large~\cite{devlin2018bert}. Specifically, we utilize the initial 19 layers of BERT-Large as the text encoder, with the subsequent 5 layers equipped with cross-attention layers serving as the multimodal decoder.

For pretraining objectives, we establish alignment across different modalities via text, including video, audio, image, and speech. We employ crossmodal contrastive and matching losses with masked language reconstruction loss as:
\begin{equation}
    \mathcal{L} = \mathcal{L}_\text{CON} + \mathcal{L}_\text{MAC} + \mathcal{L}_\text{MLM},
\end{equation}
The employed $\mathcal{L}_\text{MAC}$ and $\mathcal{L}_\text{MLM}$ are standard loss from \cite{Cheng2022VindLUAR}. Specifically, the crossmodal contrastive learning is given as:

\noindent\resizebox{\linewidth}{!}{%
\begin{minipage}{\linewidth}
\begin{align}
\resizebox{1\linewidth}{!}{
$\mathcal{L}_{\text{CON}} = \sum_{M, T_{M'}}\mathcal{L}_{\text{CON}}(M, T_{M'}) = -\sum_{M, T_{M'}} \left( \sum^N_{i=1} \log \frac{\exp(\text{sim}(f^{M}_i, f^{T_{M'}}_i) / \tau)}{\sum^N_{j=1} \exp(\text{sim}(f^{M}_i, f^{T_{M'}}_j) / \tau)} + \sum^N_{i=1} \log \frac{\exp(\text{sim}(f^{T_{M'}}_i, f^{M}_i) / \tau)}{\sum^N_{j=1} \exp(\text{sim}(f^{T_{M'}}_i, f^{M}_j) / \tau)} \right)$,}
\end{align}
\end{minipage}
}
\label{eq:loss_contrastive}
where $f^{V}$ and $f^{T}$ denote the learned video and text embeddings, respectively. $M$ and $T_{M'}$ indicates the modality of input signals and the text descriptions describing, respectively.
$\text{sim}(\cdot)$ computes the cosine similarity between two features. $\tau$ is the learnable temperature.

For the matching part, it is given as:
\begin{equation}
    \mathcal{L}_{\text{MAC}} = - y \log f_p(\mathbf{V}, \mathbf{T}) - (1-y)\log (1-f_p(\mathbf{V}, \mathbf{T})),
\end{equation}
where $f_p(\mathbf{V}, \mathbf{T})$ computes the matching likelihood between $\mathbf{V}$ and $\mathbf{T}$. $y$ denotes whether the given video and text are paired ($y=1$) or not ($y=0$).

The employed masked language modeling loss is:
\begin{equation}
    \mathcal{L}_{\text{MLM}} = - \log f_p^T(\mathbf{T}_j|\mathbf{T}_{<j}),
\end{equation}
where $f_p^T(\mathbf{T}_j|\mathbf{T}_{<j})$ computes the likelihood of the $j_\text{th}$ text token based on the previous ones. Here $\mathbf{T}$ refers to video captions.

To improve the training efficiency, we employ the masked learning strategy, aligning unmasked video tokens to tokens from other modalities first, then using full video tokens reconstruction shortly. Specifically, it consists of two steps as follows:

\paragraph{Aligning Masked Visual-Language-Audio.}
We freeze the audio encoder and focus on aligning visual, audio, and text features. For pre-training, we use a comprehensive set of image, video, and audio-video data. The combinations of modalities used are represented as $\{M, T_{M'}\} \in \{ \{I, T_I\}, \{V, T_V\}, \{V, T_\textit{VAS}\}, \{\textit{VA}, T_\textit{VAS}\}\}$ where each pair denotes the concatenated features from the respective modalities.

\paragraph{Unmasked Visual-Audio-Language Post-Pretraining.}
We freeze the vision encoder to jointly align audio, visual, and text features. Post-pretraining is conducted using a smaller subset of image and video data (25M samples), along with the full set of audio (0.5M samples) and audio-video data (50M samples). Since the parameters of the largest ViT-6B model are frozen, we do not use masking strategies in this phase to ensure consistency with the inference process and to minimize any performance degradation in downstream tasks. The modality combinations used here are $\{M, T_{M'}\} \in \{ \{I, T_I\}, \{V, T_V\}, \{A, T_{A}\}, \{V, T_\textit{VAS}\}, \{\textit{VA}, T_\textit{VA}\}\}\}$.

\subsection{Stage3: Predicting Next Token with Video-Centric Inputs}
To further enrich the semantics embedded in {\modelname} and improve its support for video-centric dialogue, we tune it by connecting it to a LLM with QFormer design~\cite{blip2,blip}. We employ the progressive learning scheme in \cite{li2023mvbench} by using \modelname{} as the video encoder and train a video blip for communicating with open-sourced LLM~\cite{vicuna,mistral}. 
Additionally, we implement a high-definition post-training stage to improve the model's fine-grained and long spatiotemporal capabilities. During this stage, the input video is divided into up to six sub-videos with a resolution of 224x224 pixels each, along with one global resized sub-video of the same resolution. We then train the model for two additional epochs: the first epoch uses 8-frame video inputs, while the second epoch uses 16-frame inputs.
During the additional training process, we update the video encoder and BLIP Qformer, while the LLM is updated using LoRA \cite{lora}.

\begin{table}[t]
\centering
\caption{Summary of datasets used in \modelname{} pretraining process.}
\resizebox{0.8\textwidth}{!}
{
\begin{tabular}{cllcl}
\toprule
Pretraining Stage & {Dataset}                         & Domain            & \# of clips & Annotation            \\ \midrule
\multicolumn{1}{c|}{Stage 1} & KMash & Web Video & 2M & - \\
\midrule
\multicolumn{1}{c|}{Stage 2 (img-txt)} & LAION, etc & Web Image & 300M & Alt-text / Generated Caps \\
\hline
\multicolumn{1}{c|}{\multirow{4}{*}{Stage 2 (vid-txt)}} & WebVid2M     & Web Video         & 250k        & Alt-text               \\
\multicolumn{1}{c|}{}                              & WebVid10M    & Web Video         & 9.7M        & Alt-text               \\
\multicolumn{1}{c|}{}                              & InternVid    & Youtube Video     & 40M         & Generated Caption     \\
\multicolumn{1}{c|}{}                              & InternVid2 & Youtube Video     & 50M         & Generated Caption     \\ 
\midrule
\multicolumn{1}{c|}{Stage 3} & LLaVA, etc  & Web Image/Video         & 2.1M        & Conversation, QA  \\
\bottomrule
\end{tabular}
}
\label{tab:pretrain_data}
\end{table}
\vspace{-2mm}
\section{Multimodal Video Data}
\vspace{-3mm}
We list our training data in Tab. \ref{tab:pretrain_data}. Among the datasets used, KMash and InternVid2 are newly built, whereas the rest are publicly available.

\subsection{Video-only Data for Masked Autoencoders}
We curate a new video set without labels named \textit{K-Mash} from action recognition datasets~\cite{k400,sth,mit,activitynet,hacs}, as detailed in Supp. It encompasses a wide range of video types, including first- and third-person perspectives, with both short and long duration, and featuring various settings. Further, we give K-Mash$_{2M}$ with additionally sourced and selected 844K videos from YouTube for diversity. 

\begin{figure*}[t]
    \centering
    \includegraphics[width=0.9\textwidth]{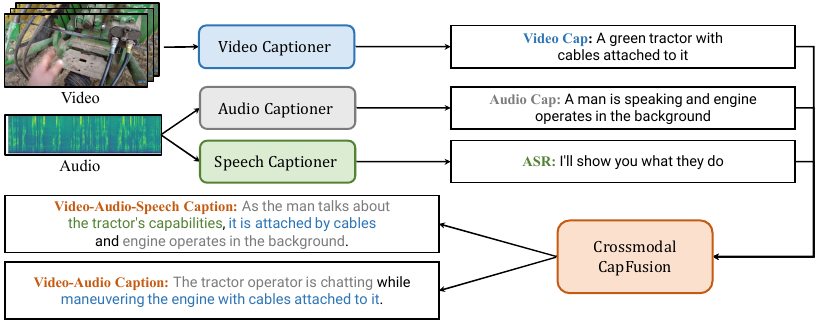}
    \caption{The framework of our video multimodal annotation system, called {\annoname}, consists of four main components: video, audio, and speech captioners, along with a LLM for integrating captions from these modalities.}
    \label{fig:anno}
\end{figure*}

\subsection{Videos with Audio-Speech Modalities}

We build a multimodal video dataset, coined as~{\dataname}, with video-audio-speech information and their descriptions for strengthening video perception via other modalities. It consists of 100M videos along with their VAS captions. 
In {\dataname}, we collect videos from several sources (detailed in the supp), segment them into clips, and automatically annotate them based on their unimodal or crossmodal inputs. We highlight the importance of temporal segmentation in clip generation and our video mutimodal annotation system. We find refining them leads to notable downstream improvements.

\paragraph{Temporal Consistency Matters.} We employ a temporal boundary detection model AutoShot~\cite{zhu2023autoshot} to segment videos into clips instead of SceneDet filter from FFmpeg. It predicts  boundaries based on temporal semantic variations rather than pixel differences, and is able to generate semantically complete cuts without mixing extra frames with inconsistent context. 

\paragraph{Video Multimodal Annotation.} We design a video multimodal annotation system {\annoname} to give proper unimodal and crossmodal descriptions for textualizing videos from different perceptions. It automatically captions visual, audio, and speech of {\dataname}, then it corrects them and fuses them for cross-modal captions via LLM. The system frame is given in Fig. \ref{fig:anno}. {\annoname} has independent video, audio, speech captioner, and a LLM for caption refinement and fusion. For video, speech, and caption postprocessing, we employ existing methods as the video captioning pipeline in \cite{wang2023internvid}, WhisperV2-large model~\cite{whisper}, and Vicuna-1.5~\cite{vicuna}. For audio, we craft a audio captioner upon VideoChat~\cite{li2023videochat}, as we find no open-sourced one. It extracts audio features from inputs by Beats~\cite{beats}. We learn it by only tuning its Qformer using a combination of the large-scale audio-text corpus WavCaps~\cite{mei2023wavcaps} dataset. Details are given in the supplementary material. 

\subsection{Instruction-Tuning Data for Video Dialogue}
We employ a updated training version of MVBench \cite{li2023mvbench}. Originally, it comprises 1.9M samples (both images and videos) from 34 distinct sources. We decrease the amount of caption data from WebVid and CoCo to 80k and 100k separately and add new data from the S-MiT to boost the diversity rather than quantity of the instruction dataset. In the additional HD training stage, we incorporate the videos with corresponding GPT-4 annotations from \cite{chen2024sharegpt4video} into the training set. We further expand the training set by including datasets from PerceptionTestQA \cite{perceptiontest}, TVQA \cite{tvqa}, NTU-RGB-D \cite{ntu_rgbd}, and EgotaskQA \cite{ego4d}, along with grounding datasets based on DiDeMo \cite{didemo} and COCO \cite{coco}.
This training data encompasses key features of image and video understanding across crucial tasks, including 1) conversation, 2) caption, 3) visual question answer, 4) reasoning, and 5) classification.

\section{Experiments} \label{sec:exp}
In our evaluation of {\modelname}, we assess the models from three learning stages. The evaluation covers a wide range of tasks, including video recognition, video retrieval, question-answering, and more. It includes various scenarios such as zero-shot learning, finetuning, and linear probing. 

For {\modelname} trained in stage 1, 2, and 3, we denote them with {\modelname}$_{s1}$, {\modelname}$_{s2}$, and {\modelname}$_{s3}$, respectively. We also learn a CLIP-style {\modelname} indicated by {\modelname}$_{clip}$. It is post-pretrained from {\modelname}$_{s2}$ by only preserving video and text encoders and contrastive loss.

Each training stage of {\modelname}-6B uses different configurations and resources. In the first stage, we employ 256 NVIDIA A100 GPUs and train the model for 18 days. The second stage also utilizes 256 A100 GPUs and spans a training period of 14 days. Finally, in the third stage, we use 64 A100 GPUs and train the model for 3 days. We introduce DeepSpeed and FlashAttention \cite{dao2022flashattention} for training and inference.
More implementation details and experiment results are given in the supp.

\subsection{Video Classification}

\subsubsection{Action Recognition} \label{sec:cls}
We test {\modelname} on Kinetics (\textit{i.e.}, K400, 600, and 700 \cite{k400,k600,k700}), Moments in Time V1 (MiT) \cite{mit}, Something-Something V2 (SSv2) \cite{sth}, UCF~\cite{soomro2012ucf101}, HMDB~\cite{hmdb}, Charades~\cite{charades_sta}, ActivityNet~\cite{activitynet} (ANet) and HACS \cite{hacs}. We evaluate in four settings: 
\emph{(a) end-to-end finetuning} the whole backbone;
\emph{(b) attentive probing} is similar to linear pooling, but extra trains the attention pooling layer~\cite{coca}. 
\emph{(c) linear probing} which freezes the backbone and only trains the task head; and
\emph{(d) zero-shot}.

\begin{table}[!t]
\caption{End-to-end finetuning action recognition results (top-1 accuracy) on Kinetics, SomethingSomething, and Moments in Time. $^{\dagger}$ denotes that the result is achieved with different resolutions or frame rates.}
\centering
\small
\resizebox{1.0\linewidth}{!}{
\setlength{\tabcolsep}{3.0 mm}{
  \begin{tabular}{llcccccccc}
    \toprule
     Method & Training Data & Setting & K400 & K600 & K700 & Sth-Sthv2 & MiT & ANet & HACS \\
    \midrule
     CoVeR~\cite{cover} & \textbf{IV}-3B & 16 $\times$ 448 & 87.1 & 87.9 & 79.8 & 70.8 & 46.1 & - & - \\
     Hiera-H~\cite{ryali2023hiera} & \textbf{V}-0.25M & 16 $\times$ 224 & 87.8 & 88.8 & 81.1 & 76.5 & -& - & - \\
     CoCa-g~\cite{coca} & \textbf{I}-3B & 16 $\times$ 576 & 88.9 & 89.4 & 82.7 & - & 49.0& - & - \\
     MTV-H~\cite{wang2023videomae} &\textbf{IV}-370M & 32 $\times$ 280 & 89.9 & 90.3 & 83.4 & - & -& - & - \\
     VideoMAEv2-g~\cite{wang2023videomae} & \textbf{V}-1.35M & 64 $\times$ 266 & 90.0 & 89.9 & - & 77.0$^{\dagger}$ & -& - & - \\
     V-JEPA-H~\cite{bardes2024vjepa} & \textbf{V}-2M & 16 $\times$ 224 & - & - & - & 77.0 & -& - & - \\
     MVD-H~\cite{wang2023masked} & \textbf{IV}-1.25M &16 $\times$ 224 & - & - & - & 77.3 & -& - & - \\
     UniFormerV2-L~\cite{uniformerv2} & \textbf{IV}-401M & 64 $\times$ 336 & 90.0 & 90.1 & 82.7 & 73.0$^{\dagger}$ & 47.8$^{\dagger}$& 94.7 & 95.4 \\
     \gray{InternVideo~\cite{wang2022internvideo}} & \textbf{V}-12M &\gray{\textit{ensemble}} & \gray{91.1} & \gray{91.3} & \gray{84.0} & \gray{77.2} & -& - & - \\
     \hline
     \modelname{}$_{s1}$-1B & \textbf{IV}-1.1M  & 8 $\times$ 224 & 91.3 & 91.4 & 85.0 & 77.1 & 50.8 & - & - \\
     \modelname{}$_{s1}$-1B & \textbf{IV}-1.1M & 16 $\times$ 224 & 91.6 & 91.6 & 85.4 & 77.1 & 50.9 & - & - \\
     \modelname{}$_{s1}$-6B  & \textbf{IV}-2M & 8 $\times$ 224 & 91.9 & 91.7 & 85.7 & \textbf{77.5} & 51.0 & - & - \\
     \modelname{}$_{s1}$-6B  & \textbf{IV}-2M & 16 $\times$ 224 & \textbf{92.1} & \textbf{91.9} & \textbf{85.9} & 77.4 & \textbf{51.2} & \textbf{95.9} & \textbf{97.0} \\
    \bottomrule
  \end{tabular}}}
\label{tab:ft_recognition1}
\end{table}

\paragraph{End-to-end Finetuning.}
Tab. \ref{tab:ft_recognition1} shows \modelname-6B obtains new state-of-the-art (SOTA) results on Kinetics (\textbf{92.1\%/91.9\%/85.9\%} on K400/600/700, respectively), SthSthv2, MiT, ANet, and HACS with only 16 frames, while the previous SOTAs require larger resolution (224 \textit{vs.} 576) or model ensemble. As for MiT in Tab. \ref{tab:ft_recognition1},
\modelname-6B exceeds the previous SOTA, CoCa-g, by a significant margin of 2.2\% (\textbf{51.2\%} \textit{vs.} 49.0\%). 
Regarding the temporal-related actions in Tab. \ref{tab:ft_recognition1},
our \modelname-6B also surpasses MVD~\cite{wang2023masked} on SSv2 (\textbf{77.5\%} \textit{vs.} 77.3\%). 
Moreover, our \modelname-6B showcases top performance on untrimmed video analysis, as indicated in Tab. \ref{tab:ft_recognition1}, 
with 95.9\% on ActivityNet and 97.0\% on HACS.
These results affirm our model's superior capability for robustly identifying complex actions across varied scenes. Note ``I" and ``V" denotes images and videos, respectively. ``IV-3B" means the total number of the used images and videos is 3B, while ``I-3B" means using 3B images.

\begin{table}[!t]

\caption{Attentive probing recognition results (top-1 accuracy) on Kinetics-400/600/700, Moments in Time and Something-Something V2.}

\centering
\resizebox{0.9\linewidth}{!}{
    \setlength{\tabcolsep}{3.0 mm}{
  \begin{tabular}{llllllll}
    \toprule
     Method& Training Data &Setting & K400 & K600 & K700 & MiT & SSV2\\
    \midrule
    UMT-L~\cite{umt} & \textbf{IV}-25M &- & 82.8  & - & - & 40.3 & 54.5 \\
    VideoMAEv2-g~\cite{wang2023videomae} & \textbf{V}-1.35M & - & 82.1  & - & - & 35.0 & 56.1 \\
    V-JEPA-H~\cite{bardes2024vjepa} & \textbf{V}-2M & 16 $\times$ 384 & 81.9  & - & - & - & \textbf{72.2} \\
    DINOv2-g~\cite{oquab2023dinov2} & \textbf{I}-142M  & 16 $\times$ 224 & 83.4  & - & - & - & 50.0 \\
    VideoPrism-g~\cite{videoprism} & \textbf{V}-619M & 16 $\times$ 288 & 87.2 & - & - & 45.5 & 68.5 \\
    ViT-e~\cite{Dehghani2023ScalingVT} & \textbf{I}-4B  & 128 $\times$ 224 & 86.5 & - & - & 43.6 & - \\
    ViT-22B~\cite{Dehghani2023ScalingVT}  & \textbf{I}-4B & 128 $\times$ 224 & 88.0 & - & - & 44.9 & - \\
    CoCa-g~\cite{coca} & \textbf{I}-3B & 16 $\times$ 576 & 88.0 & 88.5 & \textbf{81.1} & 47.4 & - \\
     \hline
     \modelname$_{s2}$-1B & \textbf{IV}-25.5M & 16 $\times$ 224 & 87.9 & 88.0 & 79.5 & 46.3 & 67.3 \\
     \modelname$_{s2}$-6B  & \textbf{IV}-400M & 16 $\times$ 224 & \textbf{88.8} & \textbf{89.1} & 81.0 & \textbf{47.8} & 67.7 \\
    \bottomrule
  \end{tabular}
}
}
\label{tab:ap_recognition}

\end{table}

\paragraph{Attentive Probing.}
As in Tab. \ref{tab:ap_recognition}, \modelname-6B not only outperforms ViT-22B \cite{Dehghani2023ScalingVT} and CoCa-g \cite{coca} in scene-focused datasets but also surpasses or matches the performance of the latest video foundation model~\cite{bardes2024vjepa,videoprism}, on datasets emphasizing temporal dynamics (SthSthV2). 
This underscores our model's exceptional ability to understand and interpret both spatial and temporal information effectively.

\begin{table}[!t]
\caption{Linear probing action recognition results (top-1 accuracy) on Kinetics-400,  Something-Something V2, UCF-101 and HMDB-51.}
\centering
\small
\resizebox{0.82\linewidth}{!}{
    \setlength{\tabcolsep}{3.0 mm}{
  \begin{tabular}{llllll}
    \toprule
     Method& Setting & K400 & SSV2 & UCF-101 & HMDB-51\\
    \midrule
    VideoMAEv2-H~\cite{wang2023videomae} & 12 $\times$ 224 & 25.8 & - & 56.4 & 34.1 \\
    TVTSv2-H~\cite{zeng2023tvtsv2} & 12 $\times$ 224 & 73.1 & - & 91.8 & 65.7 \\
    OpenCLIP-G~\cite{openclip} & 8 $\times$ 224 & 78.3 & 35.8 & 90.7 & - \\
    DINOv2-g~\cite{oquab2023dinov2} & 8 $\times$ 224 & 78.4 & 38.3 & 91.2 & - \\
     \hline
     \modelname$_{s1}$-1B & 16 $\times$ 224 & 81.6 & 46.3 & 96.0 & 71.6 \\
     \modelname$_{s1}$-6B & 16 $\times$ 224 & 82.0 & 47.8 & 96.3 & 71.8 \\
     \modelname$_{s2}$-6B & 16 $\times$ 224 & \textbf{84.2} & \textbf{56.7} & \textbf{97.3} & \textbf{80.7} \\
    \bottomrule
  \end{tabular}
}
}
\label{tab:lp_recognition}
\end{table}

\paragraph{Linear Probing.}
In Tab. \ref{tab:lp_recognition}, \modelname-1B significantly outperforms the previous SOTA, DINOv2-g~\cite{oquab2023dinov2}, by notable margins: +3.2\% on K400, +8.0\% on SthSthV2, and +4.8\% on UCF-101. As we scale the model, an upward trend in results is observed, underscoring the benefits of model enhancement. Notably, the integration of multimodal pretraining (stage 2) yields further rise in results. We suppose stage 2 enhances feature discrimination.

\begin{table}[!t]
\centering
\caption{Zero-shot action recognition on UCF, HMDB, MiTv1, SSv2-MC, and Charades.}
{
\resizebox{0.78\textwidth}{!}{
\setlength{\tabcolsep}{1.0 mm}{
  \begin{tabular}{lrllllll}
    \toprule
     {Method} & \#F & {Training Data} &{UCF} & HMDB & MiT & SSv2-MC & Charades\\
     \midrule
     CLIP~\cite{clip} & 12&  \textbf{I}-400M  & 68.9 & 43.2 & - & 29.6 & - \\ 
     TVTSv2~\cite{zeng2023tvtsv2} & 12 & \textbf{V}-8.5M & 78.0 & 52.1 & - & 48.4 & - \\
     VideoCoCa-g~\cite{videococa} & 16 & \textbf{V}-145M & - & - & - & - & 25.8 \\
     VideoPrism-g~\cite{videoprism} & 16 & \textbf{V}-619M & - & - & - & - & 32.4 \\
    \midrule
     {\modelname}$_{clip}$-1B & 8 & \textbf{IV}-25.5M & 88.8 & 53.9 & 31.6 & 61.5 & 32.9 \\
     {\modelname}$_{clip}$-6B & 8 & \textbf{IV}-400M & \textbf{89.5} & \textbf{56.7} & \textbf{32.9} & \textbf{63.5} & \textbf{34.6} \\
    \bottomrule
  \end{tabular}
  }
}
}
\label{tab:zs_recognition2}
\end{table}
\begin{table}[!t]
\caption{Zero-shot action recognition results on Kinetics. }
\centering
\resizebox{0.6\textwidth}{!}{
\small
\setlength{\tabcolsep}{1.0 mm}{
  \begin{tabular}{lrlllllll}
    \toprule
     \multirow{2}{*}{Method}& \multirow{2}{*}{\#F} &\multicolumn{2}{c}{K400} & \multicolumn{2}{c}{K600} & \multicolumn{2}{c}{K700}\\
     & & top-1 & AVG & top-1 & AVG  &top-1  & AVG \\
    \midrule
     CLIP~\cite{clip} & 8 & 58.4 & 70.1 & 55.1 & 67.2 & 46.1 & 58.4 \\
     EVA-CLIP-L~\cite{eva_clip} & 1 & - & 65.0 & - & 64.9 & - & 59.1 \\
     EVA-CLIP-E~\cite{eva_clip} & 1  & - & 69.8 & - & 69.3 & - & 63.4 \\
    ViCLIP-L~\cite{wang2023internvid} & 8  & 64.8 & 75.7 & 62.2 & 73.5 & 54.3 & 66.4\\
    VideoCoCa-g~\cite{videococa} & 16  & 72.0 & 81.3 & - & - & - & -\\
     InternVL-6B~\cite{chen2023internvl} & 8  & 69.1 & {79.4} & {68.9} & {78.8} & {60.6} & {71.5}\\
     EVA-CLIP-18B~\cite{sun2024eva} & 16 & - & {79.4} & - & {79.4} & - & {72.2} \\
     VideoPrism-g~\cite{videoprism} & 16& \textbf{76.4} & \textbf{85.4} & - & - & - & -\\ \midrule

     {\modelname}$_{clip}$-1B & 8  & \underline{73.1} & \underline{82.4} & \textbf{72.8} & \textbf{81.8} & \textbf{64.9}  & \textbf{75.2} \\
     {\modelname}$_{clip}$-6B & 8  & 72.7 & {82.2} & 71.7 & 81.2 & 64.2 & 75.2 \\
    \bottomrule
  \end{tabular}
}
}
\label{tab:zs_recognition}
\end{table}

\paragraph{Zero-shot.} Table~\ref{tab:zs_recognition2} and \ref{tab:zs_recognition} show {\modelname} gets 72.7\% / 71.7\% / 64.2\% on K400/600/700, respectively, outperforming others but VideoPrism on K400 (76.4\%). On UCF~\cite{soomro2012ucf101}, HMDB~\cite{hmdb}, MiT~\cite{mit}, SSv2-MC, and Charades, {\modelname} gives an cutting edges over others. The clear gap between VideoPrism and {\modelname} on K400 may signify the importance of pretraining corpus in VideoPrism (311M videos with text and 36.1M of them are manually labeled) for K400 in zero-shot. Note that on the datasets of Kinetics, UCF101 and HMDB51, which have a distribution closer to the pre-training dataset used in stage1, the performance of Internvideo2-6B is slightly inferior to that of Internvideo2-1B. We suppose this is caused by Internvideo2-6B uses a more abundant pretraining dataset in stage2, leading to the forgetting of pretraining data in stage1.

\begin{table}[t]
    \caption{Finetuned temporal action localization results on THUMOS14~\cite{thumos}, ActivityNet~\cite{anet}, HACS Segment~\cite{hacs} and FineAction~\cite{liu2022fineaction}. We report average mAP. ``Flow'' denotes the ensembling I3D flow feature. * denotes the result is achieved with Flow.}
    \centering
    \resizebox{0.7\linewidth}{!}{
    \setlength{\tabcolsep}{2pt} %
    \begin{tabular}{lcccc}
        \toprule
        Backbone & THUMOS14 & HACS & ActivityNet & FineAction \\
        \midrule
        I3D \cite{k400} + Flow & 66.8 & - & 35.6 & - \\
        R(2+1)D \cite{r(2+1)d} & 55.6 & - & 36.6 & - \\
        InternVideo & 71.6$^*$ & 41.3 & 39.0 & 17.6 \\
        VideoMAEv2-g \cite{wang2023videomae} & 69.5 & - & - & 18.2 \\
        {\modelname}$_{s1}$-1B & 69.8 & 42.4 & 40.4 & 27.2 \\
        {\modelname}$_{s1}$-6B & \textbf{72.0} & \textbf{43.3} & \textbf{41.2} & \textbf{27.7}\\
        \bottomrule
    \end{tabular}
    }
     \label{tab:exp-tal}
\end{table}

\subsubsection{Temporal Action Localization} \label{sec:tal}
We evaluate models on four temporal action localization (TAL) datasets: THUMOS14~\cite{thumos}, ActivityNet~\cite{anet}, HACS Segment~\cite{hacs} and FineAction~\cite{liu2022fineaction} in a feature-based manner with finetuning. We employ output of the 7-th layer from {\modelname} for inputs as the corresponding features without fusing anything else. ActionFormer~\cite{didemo} is used as the detection head. We report mean Average Precision(mAP) under multiple tIoU as in~\cite{lin2019bmn,chen2022dcan,didemo,yang2023basictad}. In Table~\ref{tab:exp-tal}, {\modelname}-6B gets the highest mAP among all comparisons in all datasets, while {\modelname}-1B nearly surpass other methods except in THUMOS14. We find {\modelname}-6B almost consistently improves mAP with a notable margin from {\modelname}-1B except in FineAction. We suppose scaling model capacity without data refinement cannot nontrivially improve fine-grained discrimination abilities of models. Scaling detailed annotations in training may address this issue.

\begin{table}[t]
\centering
\caption{Video instance segmentation performance (mAP) on YouTube-VIS19 \cite{yang2019video}. }
\resizebox{0.6\linewidth}{!}{
  \begin{tabular}[t]{lccc}
    \toprule
     Method& Backbone & \#Params & YouTubeVIS19 \\
    \midrule
    Mask2Former & Swin-L (image) \cite{swin} & 219M & 60.3 \\
    Mask2Former & InternViT (image) & 6B & 63.4 \\
    \midrule
    Mask2Former & {\modelname}$_{s1}$ & 6B & \textbf{64.2} \\
    \bottomrule
  \end{tabular}
  }
\label{tab:videoinstanceseg}
\end{table}
\subsubsection{Video Instance Segmentation} \label{sec:vis}
We evaluate on the Video Instance Segmentation (VIS) dataset Youtube-VIS 2019~\cite{yang2019video}. Built upon Mask2Former~\cite{cheng2021maskformer}, we employ the video encoder of {\modelname} as backbone with ViT-adapter~\cite{chen2022vitadapter} for features. We also try InternViT~\cite{chen2023internvl} for comparisons. In Table~\ref{tab:videoinstanceseg}, {\modelname} gets the highest mAP among all. This validates its effectiveness in relatively fine-grained spatiotemporal perception.

\subsection{Video-Audio-Language Tasks}
We evaluate {\modelname} on video retrieval, captioning, and multi-choice question-answersing (QA). The former two tasks are conducted by matching video representation and the candidate text ones using the text encoder in stage 2. The latter is tested by the VideoLLM learned in stage 3. We also test audio tasks.

\begin{table}[!t]
\caption{Results of zero-shot video retrieval in both text-to-video (T2V) and video-to-text (V2T) on MSR-VTT, LSMDC, DiDeMo, MSVD, ActivityNet (ANet), and VATEX.}
\centering
\small
\resizebox{.9\textwidth}{!}{
\setlength{\tabcolsep}{1.5 mm}{
  \begin{tabular}{llllllllllllll}
    \toprule
    \multirow{2}{*}{Method}& \multicolumn{2}{c}{MSR-VTT} & \multicolumn{2}{c}{LSMDC} & \multicolumn{2}{c}{DiDeMo} & \multicolumn{2}{c}{MSVD} & \multicolumn{2}{c}{ANet} & \multicolumn{2}{c}{VATEX}\\
    & T2V & V2T & T2V & V2T & T2V & V2T & T2V & V2T & T2V & V2T & T2V & V2T \\
    \midrule
     CLIP~\cite{clip} & 30.4 & 24.2 & 13.9 & 11.9 & 12.7 & 18.7 & 40.5 & 57.2 & 9.1 & 13.2 &- &-\\
     CLIP4Clip~\cite{clip4clip} & 32.0 & - & 15.1 & - & - & - & 38.5 & - & - & - & -&-\\
     ViCLIP~\cite{wang2023internvid} & 42.4 & 41.3 & 20.1 & 16.9 & 18.4 & 27.9 & 49.1 & 75.1 & 15.1 & 24.0 & - & -\\
     InternVideo-L~\cite{wang2022internvideo} & 40.7 & 39.6 & 17.6 & 13.2 & 31.5 & 33.5 & 43.4 & 67.6 & 30.7 & 31.4 & 49.5 & 69.5\\
     UMT-L~\cite{umt} & 40.7 & 37.1 & 24.9 & 21.9 & 48.6 & 49.9 & 49.0 & 74.5 & 41.9 & 39.4 & - & -\\
     VideoCoCa-g~\cite{videococa}  & 34.4 & 64.7 & - & - & - & - & - & - & 34.5 & 33.0 & 53.2 & 73.6\\
     VideoPrism-g~\cite{videoprism}  & 39.7 & \textbf{71.0} & - & - & - & - & - & - & 52.7 & 50.3 & 62.5 & 77.1\\
     \midrule
     \modelname$_{s2}$-1B  & 51.9 & 50.9 & 32.0 & 27.3 & 57.0 & 54.3 & 58.1 & 83.3 & 60.4 & 54.8 & 70.4 & 85.4\\
     \modelname$_{s2}$-6B & \textbf{55.9} & 53.7 & \textbf{33.8} & \textbf{30.1} & \textbf{57.9} & \textbf{57.1} & \textbf{59.3} & \textbf{83.1} & \textbf{63.2} & \textbf{56.5} & \textbf{71.5} & \textbf{85.3}\\
    \bottomrule
  \end{tabular}
}
}
\label{tab:zs_retrieval}
\end{table}

\begin{table}[!t]
\caption{Results of finetuning video retrieval in both text-to-video (T2V) and video-to-text (V2T) on MSR-VTT, LSMDC, DiDeMo, MSVD, ActivityNet (ANet), and VATEX. }
\centering
\small
\resizebox{.9\textwidth}{!}{
\setlength{\tabcolsep}{1.5 mm}{
  \begin{tabular}{lcllllllllllll}
    \toprule
    \multirow{2}{*}{Method}& \multicolumn{2}{c}{MSR-VTT} & \multicolumn{2}{c}{LSMDC} & \multicolumn{2}{c}{DiDeMo} & \multicolumn{2}{c}{MSVD} & \multicolumn{2}{c}{ANet} & \multicolumn{2}{c}{VATEX}\\
      & T2V & V2T & T2V & V2T & T2V & V2T & T2V & V2T & T2V & V2T & T2V & V2T\\
    \midrule
    CLIP~\cite{clip} & 38.2 & 38.7 & 22.5 & 22.6 & 32.2 & 33.9 & - & - & 26.1 & 26.9 & - & - \\
     CLIP4Clip~\cite{clip4clip}  & 45.6 & 45.9 & 24.3 & 23.8 & 43.0 & 43.6 & 45.2 & 48.4 & 40.3 & 41.6& - & - \\
     ViCLIP~\cite{wang2023internvid}  & 52.5 & 51.8 & 33.0 & 32.5 & 49.4 & 50.2 & - & - & 49.8 & 48.1& - & - \\
     UMT-L~\cite{umt}  & 58.8 & 58.6 & 43.0 & 41.4 & 70.4 & 65.7 & 58.2 & 82.4 & 66.8 & 64.4 & 72.0 & 86.0 \\
     \midrule
     \modelname$_{s2}$-6B  & \textbf{62.8} & \textbf{60.2} & \textbf{46.4} & \textbf{46.7} & \textbf{74.2} & \textbf{71.9} & \textbf{61.4} & \textbf{85.2} & \textbf{74.1} & \textbf{69.7} & \textbf{75.5} & \textbf{89.3}\\
    \bottomrule
  \end{tabular}
}}
\label{tab:retrieval}
\end{table}

\subsubsection{Video Retrieval} \label{sec:retrieval}
We evaluate the video retrieval on six popular benchmarks: MSR-VTT \cite{msrvtt}, LSMDC \cite{lsmdc}, DiDeMo \cite{didemo}, MSVD \cite{msvd}, ActivityNet (ANet) \cite{activitynet}, and VATEX \cite{vatex}, as shown in Tab. \ref{tab:zs_retrieval} and \ref{tab:retrieval}. In evaluation, eight frames from the input videos are uniformly sampled. \textit{We report R@1 scores for both text-to-video (t2v) and video-to-text (v2t) tasks} in Tab. \ref{tab:zs_retrieval} and \ref{tab:retrieval}. R@5 and R@10 are given in Supp.

Tab. \ref{tab:zs_retrieval} and \ref{tab:retrieval} demonstrate that {\modelname} outperforms other state-of-the-arts with a notable margin in both t2v and v2t of all used datasets no matter in zero-shot or finetuned settings, except for the v2t of MSR-VTT, where VideoPrism gives the best result. This shows the video-language semantic alignment of transferrity of {\modelname}.

\begin{table}[!t]
    \caption{Finetuned temporal grounding on QVHighlight~\cite{lei2021qvhighlights} and Charade-STA~\cite{charades_sta}. 
    }
    \resizebox{\textwidth}{!}{
    \begin{subtable}[t]{.5\linewidth}
		\centering
		\caption{QVHighlight}
    	\setlength{\tabcolsep}{.5pt} %
		\vspace{-0.3\baselineskip}
		\scriptsize{
			\begin{tabular}{lccc|cc}
            \toprule
             Feature  & R1@0.5 & R1@0.7 & mAP & mAP & HiT@1 \\
             \midrule
             CLIP \cite{clip}  &  64.97 & 48.65 &42.96 & 39.83 & 64.19  \\
             CLIP+SlowFast \cite{slowfast} & 65.43 & 48.38 & 42.86 & 40.33 & 66.21  \\
             \modelname$_{s2}$-1B &  \underline{70.00} & \underline{54.45} & \underline{47.02} & \underline{42.36} & \underline{69.74}  \\
             \modelname$_{s2}$-6B &  \textbf{71.42} & \textbf{56.45} & \textbf{49.24} & \textbf{42.90} & \textbf{72.00}  \\
            \bottomrule
          \end{tabular}
		}
		\label{tab:grounding-qvhighlight}
	\end{subtable}
  \hfill
    \begin{subtable}[t]{.5\linewidth}
		\centering
		\caption{Charade-STA}
    	\setlength{\tabcolsep}{.5pt} %
		\vspace{-0.3\baselineskip}
		\scriptsize{
			\begin{tabular}{lcccc}
            \toprule
             Feature  & R1@0.3 &R1@0.5 & R1@0.7 & mIoU   \\
             \midrule
             CLIP \cite{clip}  & 65.62 & 52.77 &30.16 &45.85   \\
             CLIP+SlowFast \cite{slowfast} & 70.43 & 58.44 & 36.34 & 50.13  \\
             \modelname$_{s2}$-1B & \underline{78.41} & \underline{68.36} & \underline{45.03} & \underline{57.12}  \\
             \modelname$_{s2}$-6B & \textbf{79.70} & \textbf{70.03} & \textbf{48.95} & \textbf{58.79}  \\
             
            \bottomrule
          \end{tabular}

		}
	\end{subtable}
 }
\label{tab:grounding} 
\end{table}

\subsubsection{Video Temporal Grounding} \label{sec:ground}

We evaluate {\modelname} on two video temporal grounding (VTG) datasets: QVhighlight~\cite{lei2021qvhighlights}, and Charade-STA~\cite{charades_sta}. The eval setting and used features are the same as in TAL.
We use CG-DETR~\cite{moon2023cg-detr} as the grounding head. 
We report R1@0.3, R1@0.5, R1@0.7, and mAP for moment retrieval as in~\cite{lei2021qvhighlights,moon2023cg-detr,lin2023univtg}. Highlight Detection is evaluated in terms of ``Very Good'' mAP and HiT@1.
In Table~\ref{tab:grounding}, \modelname-1B and \modelname-6B bring gradual performance improvements compared to CLIP~\cite{clip} and CLIP~\cite{clip}+Slowfast~\cite{slowfast}. This suggests that a larger spatiotemporal model is more beneficial to short-term video semantic alignment capabilities.

\begin{table}[!t]
\caption{Audio retrieval results on AudioCaps \cite{kim2019audiocaps}, Clothov1, and Clothov2 \cite{drossos2020clotho}. We report text-to-audio R@1 accuracy in zero-shot and finetuning settings. }
\centering
\small
\resizebox{0.87\textwidth}{!}{
\setlength{\tabcolsep}{1.5 mm}{
    \begin{tabular}{l|ccc|ccc}
        \toprule
        \multirow{2}{*}{Method} & \multicolumn{3}{c}{Zero-shot} & \multicolumn{3}{c}{Finetuning} \\
         & AudioCaps & ClothoV1 & ClothoV2 & AudioCaps & ClothoV1 & ClothoV2 \\
         \midrule
        VIP-ANT \cite{zhao2021connecting} & 27.7 &  & - & - &  & - \\
        VAST \cite{chen2024vast} & - & - & - & 52.0 & 25.1 & 26.9 \\
        LanguageBind \cite{zhu2023languagebind} & - & 12.1 & 12.1 & - & - & - \\
        {\modelname}$_{s2}$-6B & \textbf{37.1} & \textbf{17.4} & \textbf{17.4} & \textbf{55.2} & \textbf{25.3} & \textbf{27.2}\\
        \bottomrule
    \end{tabular}
}}
\label{tab:audio_retrieval}
\end{table}

\begin{table}[!t]
    \caption{Results of AudioQA on ClothoAQA \cite{lipping2022clotho} and Audio-MusicAVQA (AMAVQA) \cite{behera2023aquallm}, and audio classification on the ESC-50 \cite{piczak2015esc}. Both in the finetuning setting. }
    \resizebox{.5\textwidth}{!}{
    \begin{subtable}[t]{.5\linewidth}
		\centering
		\caption{ClothoAQA and AMAVQA}
    	\setlength{\tabcolsep}{2pt} %
		\renewcommand*{\arraystretch}{1.10}  %
		\vspace{-0.3\baselineskip}
		\scriptsize{
			\begin{tabular}{l|cc}
                \toprule
                Backbone & ClothQA & AMAVQA \\
                \midrule
                AquaNet \cite{lipping2022clotho} & 14.78 & 65.59 \\
                MWAFM \cite{li2023multi} & 22.24 & 67.54 \\
                {\modelname}$_{s2}$ & \textbf{30.14} & \textbf{80.51} \\
                \bottomrule
            \end{tabular}
		}
		\label{tab:audioqa}
	\end{subtable}
    }
    \resizebox{.5\textwidth}{!}{
    \begin{subtable}[t]{.5\linewidth}
		\centering
		\caption{ESC-50}
    	\setlength{\tabcolsep}{2pt} %
		\renewcommand*{\arraystretch}{1.10}  %
		\vspace{-0.3\baselineskip}
		\scriptsize{
			\begin{tabular}{l|c}
            \toprule
             Method  & Top-1 Acc \\
             \midrule
             AST \cite{gong2021ast}  & 95.60    \\
             BEATs  \cite{beats}     & 98.10 \\
             {\modelname}$_{s2}$   & \textbf{98.60} \\
            \bottomrule
          \end{tabular}
        \label{tab:audiocls}
		}
        \end{subtable}
        }
\end{table}

\subsubsection{Audio-related Tasks} \label{sec:audio}
We evaluate {\modelname}'s audio and text encoders on audio tasks, including audio-text retrieval on AudioCaps \cite{kim2019audiocaps}, Clothov1, and Clothov2 \cite{drossos2020clotho}; audioQA on ClothoAQA \cite{lipping2022clotho} and Audio-MusicAVQA \cite{behera2023aquallm}; and audio classification on the ESC-50 \cite{piczak2015esc}. As shown in Tab.~\ref{tab:audio_retrieval}, ~\ref{tab:audioqa}, and \ref{tab:audiocls}, our model achieves state-of-the-art performance on all downstream tasks. Considering the limited size of the used audio and text encoders, these audio-related results show crossmodal contrastive learning's benefits are mutual to the used modalities. Audio and the corresponding text models also gain from this learning.

\begin{table*}[!t]
    \caption{
        Results of Chat-centric Evaluation on Multiple Choice Video-QA on MVBench \cite{li2023mvbench}, Egoschema~\cite{egoschema} and Perception Test~\cite{perceptiontest}. 
    }
    \centering
    \small
    \resizebox{1.0\linewidth}{!}{
    \setlength{\tabcolsep}{3.0 mm}{
        \begin{tabular}{l|l|l|c|c|c}
        \toprule
        \textbf{Model} & \textbf{ViEncoder} & \textbf{LLM} & MVBench  & Egoschema  & Perception Test \\
        \midrule
        GPT-4V~\cite{gpt4v} & - & GPT-4 & {43.5} & -   & -  \\
        Gemini 1.0 Pro~\cite{team2023gemini} & - & - & 37.7  & 55.7 & 51.1 \\
        Gemini 1.0 Ultra~\cite{team2023gemini} & - & - & - & 61.5 & 54.7 \\
        Gemini 1.5 Pro~\cite{team2023gemini} & - & - & -  & \textbf{72.2}  & - \\
        
        \midrule
        LLaVA-Next-Video \cite{liu2024llava} & CLIP-L & Vicuna-7B & 46.5 & 43.9 & 48.8  \\
        VideoLLaMA2 \cite{cheng2024videollama} & CLIP-L-336 & Mistral-7B & 54.6 & 51.7  & 51.4 \\
        VideoLLaMA2 \cite{cheng2024videollama} & CLIP-L-336 & Mistral-8*7B & 53.9 & 53.3  & 52.2 \\
        \hline
        \textbf{VideoChat2} & UMT-L & Vicuna-7B & 51.1 & -  & - \\
        \textbf{VideoChat2} & {\modelname}$_{s3}$-1B & Mistral-7B & 60.3 & 55.8  & 53.0 \\
        \textbf{VideoChat2-HD} & {\modelname}$_{s3}$-1B & Mistral-7B & 65.4 & 60.2  & 60.1 \\
        \textbf{VideoChat2-HD-F16} & {\modelname}$_{s3}$-1B & Mistral-7B & \textbf{67.2} & 60.0 & \textbf{63.4} \\
        \bottomrule
        \end{tabular}
        }
    }
    \label{tab:wct-vcbench}
\end{table*}

\begin{table}[!t]
\caption{Average top-1 accuracy of action recognition (K400, SSv2, and MiT) and video retrieval (MSR-VTT, LSMDC, DiDeMo, MSVD, ANet, and VATEX in t2v) using zero-shot and finetuning settings. * denotes results are from {\modelname}$_{s1}$.}
\centering
\small
\resizebox{0.7\textwidth}{!}{
\setlength{\tabcolsep}{1.5 mm}{
    \begin{tabular}{l|ccc}
        \toprule
        \multirow{2}{*}{Model} & \multicolumn{2}{c}{Zero-shot}  & Finetuning \\
         & Action Recognition & Video Retrieval & Action Recognition \\
         \midrule
        {\modelname}$_{s2}$-1B & 55.5 & 55.0 & 73.2* \\
        {\modelname}$_{s2}$-6B & \textbf{56.9}$_{(+1.4)}$ & \textbf{56.9}$_{(+1.9)}$ & \textbf{73.6}$_{(+0.4)}$* \\
        \bottomrule
    \end{tabular}
    }}
\label{tab:model_scale}
\end{table}

\begin{table}[!t]
\vspace{-0mm}
\caption{Ablation of Stage1, conducted with finetuned action recognition on Kinetics, MiT, and SthSthv2. All models are tested with 8$\times$224$\times$224 input.}
\centering
\small
\resizebox{1.0\linewidth}{!}{
\setlength{\tabcolsep}{3.0 mm}{
  \begin{tabular}{lllllllll}
    \toprule
     Model & Teacher & Data & K400 & K600 & K700 & MiT & SSv2 & Avg \\
    \midrule
    ViT-L & CLIP-L & K710 & 90.3 & 90.4 & 83.2 & 48.0 & 74.7 & 77.3 \\
    ViT-L & CLIP-L & K-Mash$_{1.1M}$ & 90.5 & 90.4 & 83.4 & 48.1 & 74.7 & 77.4 \\
    \hline
    ViT-1B & InternVL-6B & K710 & 90.9 & 91.0 & 84.7 & 49.8 & 75.9 & 78.5 \\
    ViT-1B & InternVL-6B & K-Mash$_{1.1M}$ & 91.4 & 91.5 & 85.1 & 50.5 & 76.5 & 79.0 \\
    ViT-1B & InternVL-6B$+$VideoMAE-g & K-Mash$_{1.1M}$ & 91.3 & 91.4 & 85.0 & 50.8 & 77.1 & 79.1 \\
    ViT-1B & InternVL-6B & K-Mash$_{2M}$ & 91.3 & 91.5 & 85.1 & 50.6 & 76.6 & 79.0 \\ 
    \hline
    ViT-6B & InternVL-6B$+$VideoMAE-g & K-Mash$_{2M}$ & \textbf{91.9} & \textbf{91.7} & \textbf{85.7} & \textbf{51.0} & \textbf{77.5} & \textbf{79.6} \\
    \bottomrule
  \end{tabular}
}
}
\label{tab:abaltion_stage1}
\end{table}

\begin{figure*}[!t]
    \centering
    \includegraphics[width=0.9\textwidth]{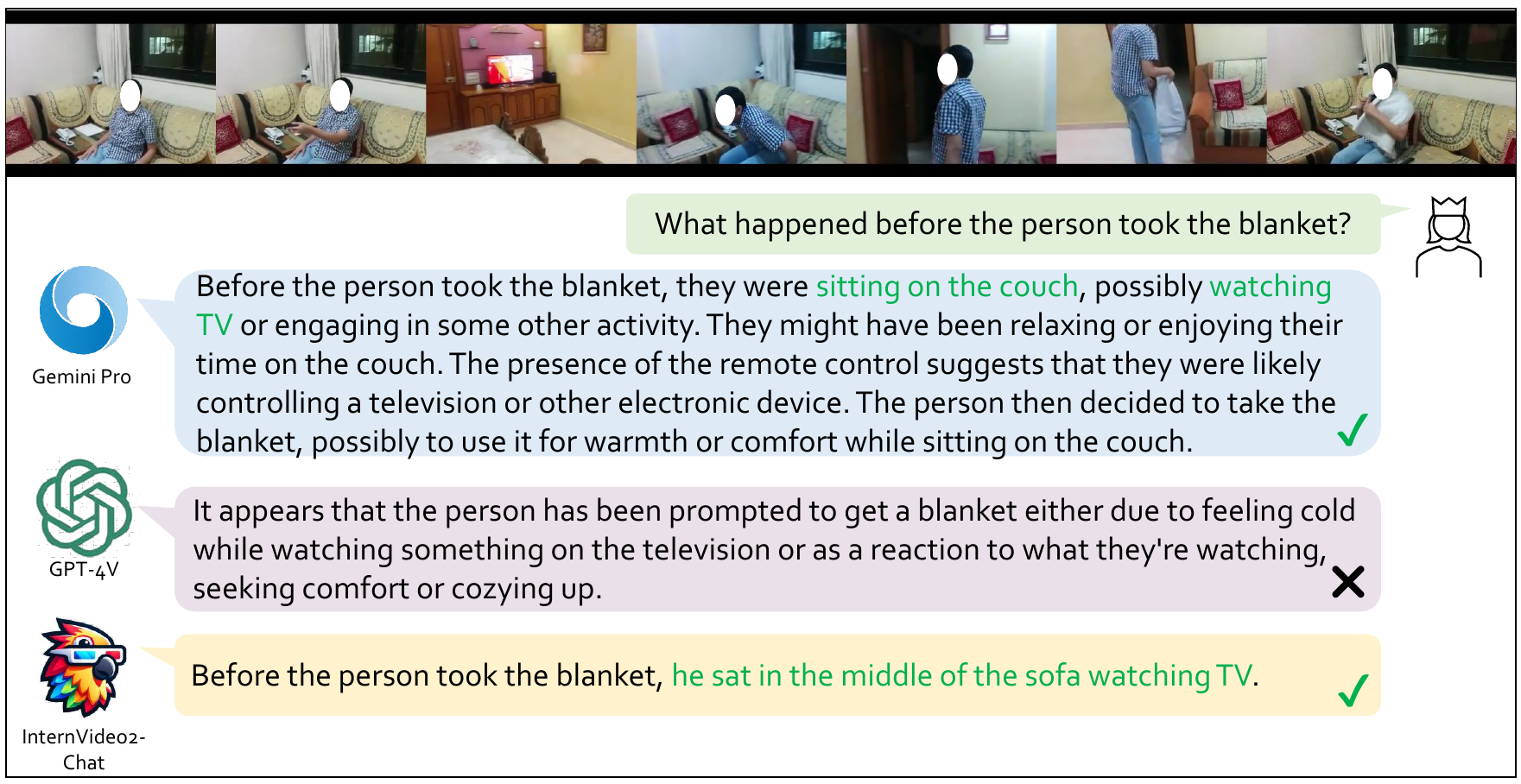}
    \caption{\textbf{Temporal action recognition tasks.} In questions about an action before it happens, Gemini Pro and InternVideo2-Chat both describe accurately the action, while GPT-4V hallucinates.}
    \label{fig:chat_action}
\end{figure*}

\begin{figure*}[!t]
    \centering
    \includegraphics[width=0.9\textwidth]{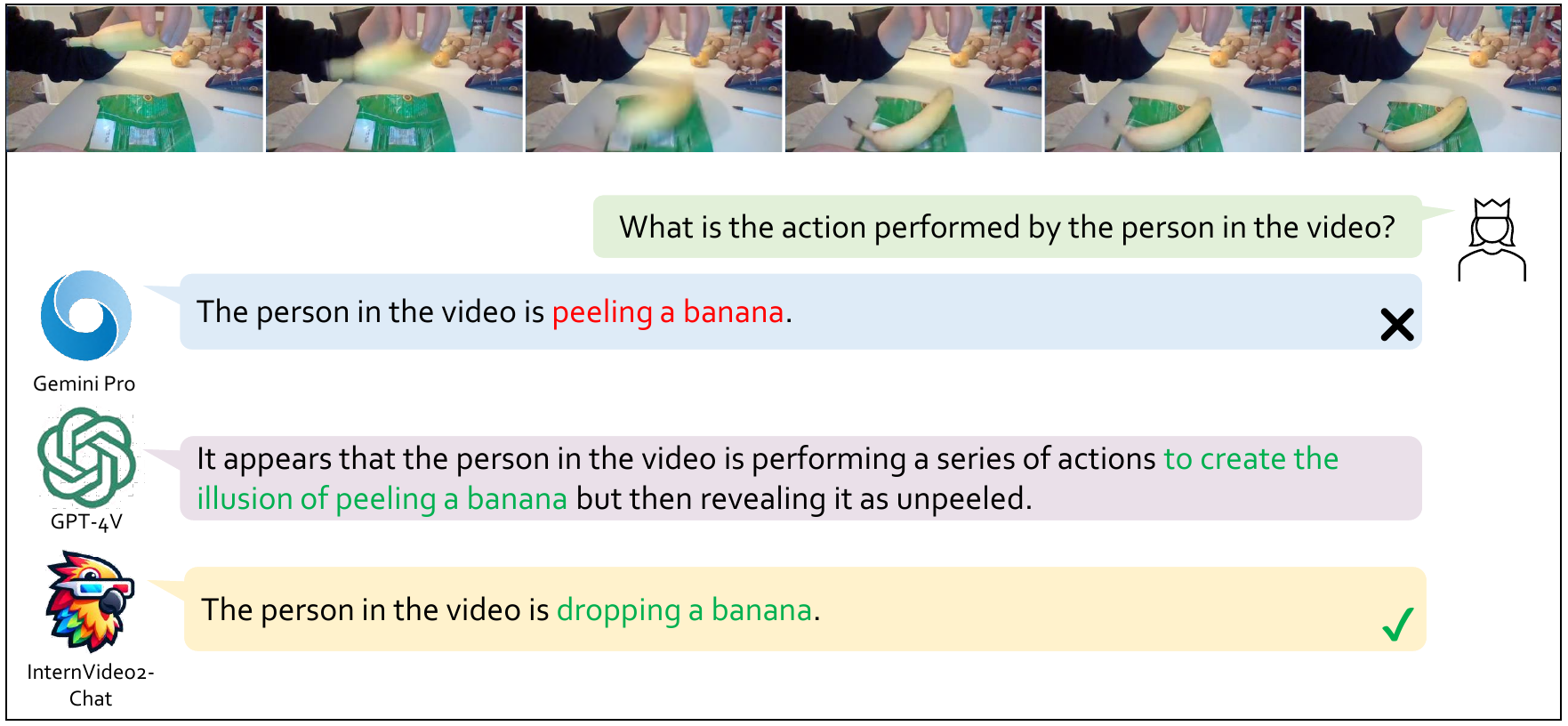}
    \caption{\textbf{Confused action recognition.} The person in the video is performing a misleading action while holding a banana. Gemini Pro gives a wrong answer. GPT-4V identifies the misleading action but doesn't give a correct answer. InternVideo2-Chat gives a correct answer.}
    \label{fig:chat_fine}
\end{figure*}

\begin{figure*}[!t]
    \centering
    \includegraphics[width=0.9\textwidth]{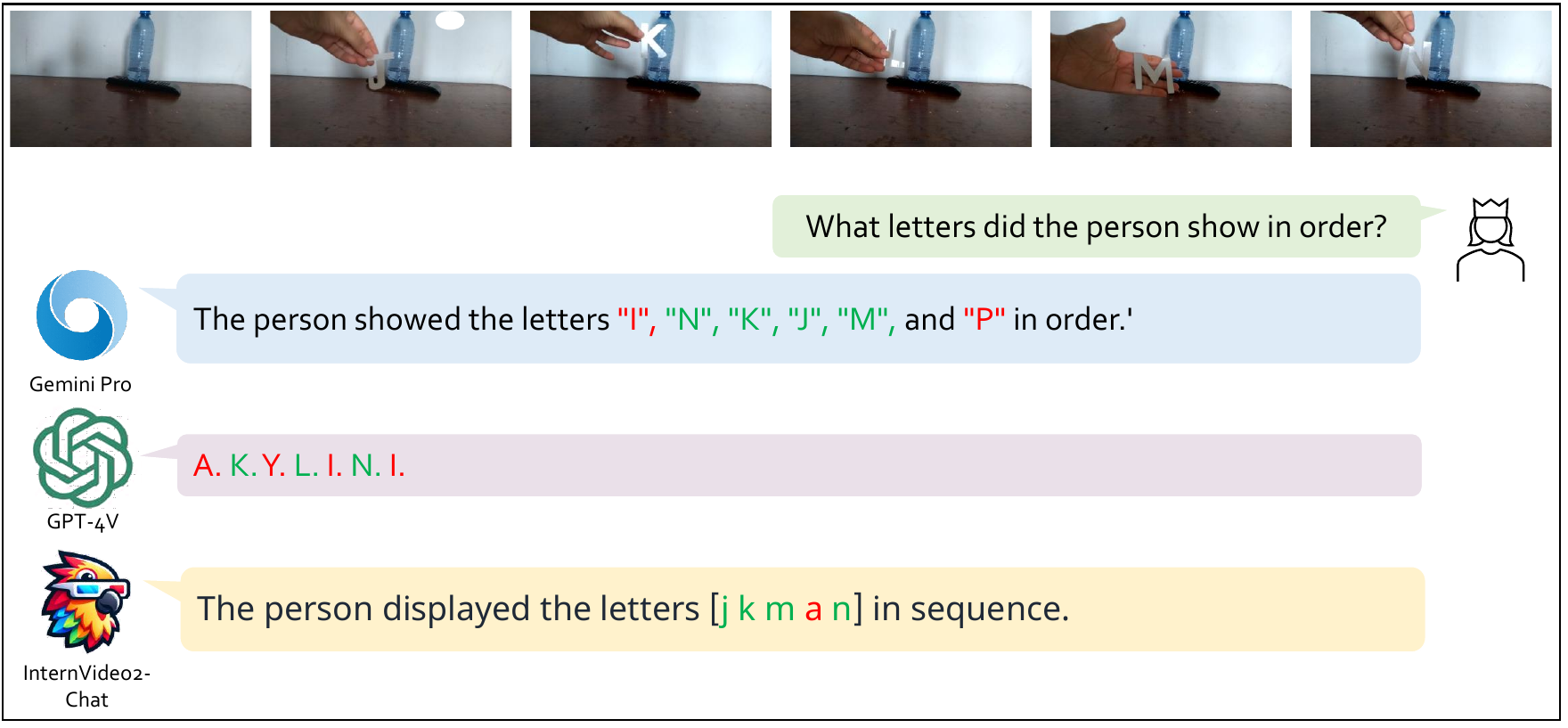}
    \caption{\textbf{Video Object Temporal Recognition.} The person in this video takes out different letters in the order of time. Gemini Pro recognizes 4 letters, but the order is totally reversed; GPT-4V recognizes only 3 letters, and the result is mixed with wrong answers; InternVideo2-Chat has the fewest errors among them and the order is correct.}
    \label{fig:chat_order}
\end{figure*}

\begin{figure*}[!t]
    \centering
    \includegraphics[width=0.9\textwidth]{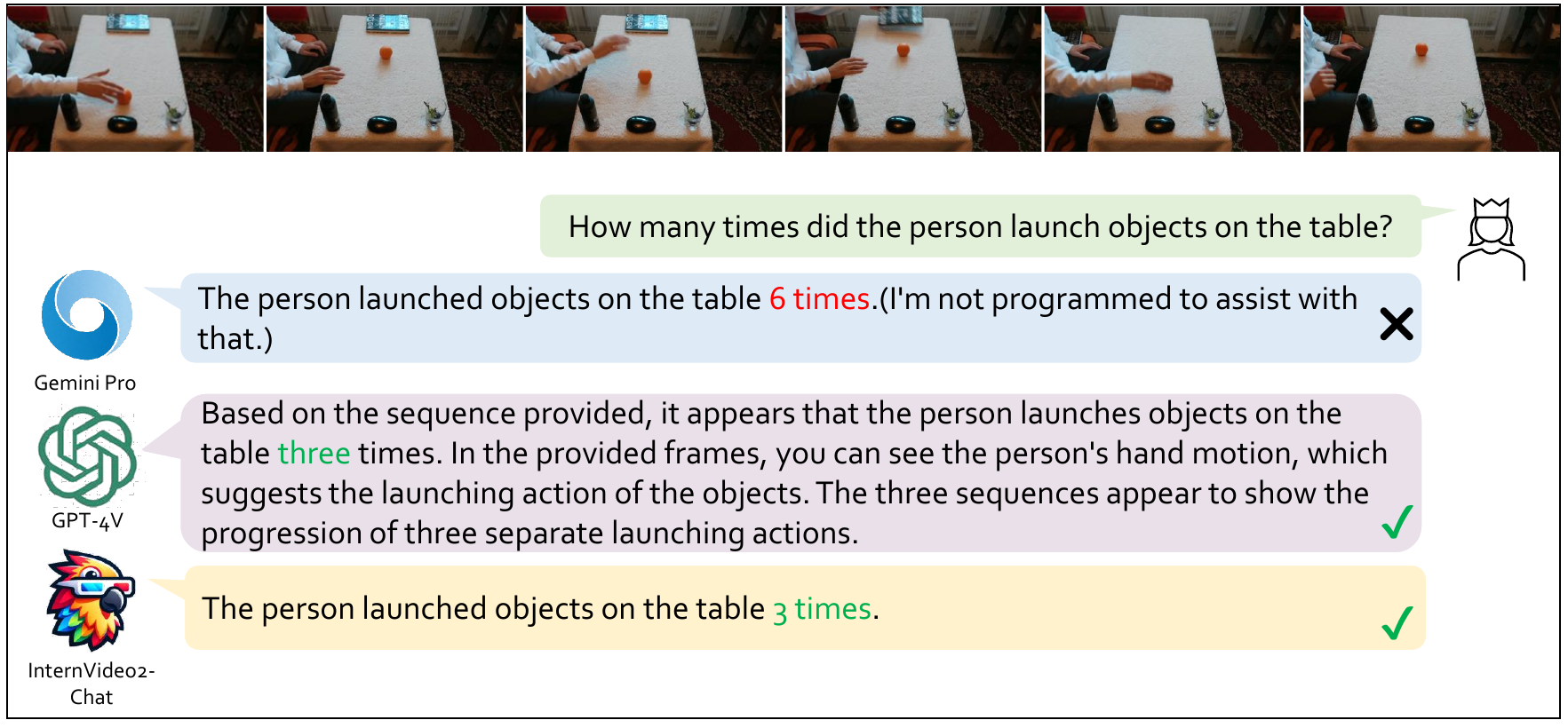}
    \caption{\textbf{Event counting task.} Both GPT-4V and InternVideo2-Chat are able to correctly capture the times of actions and not be confused by redundant frames and other actions.}
    \label{fig:chat_count}
\end{figure*}

\begin{figure*}[!t]
    \centering
    \includegraphics[width=0.9\textwidth]{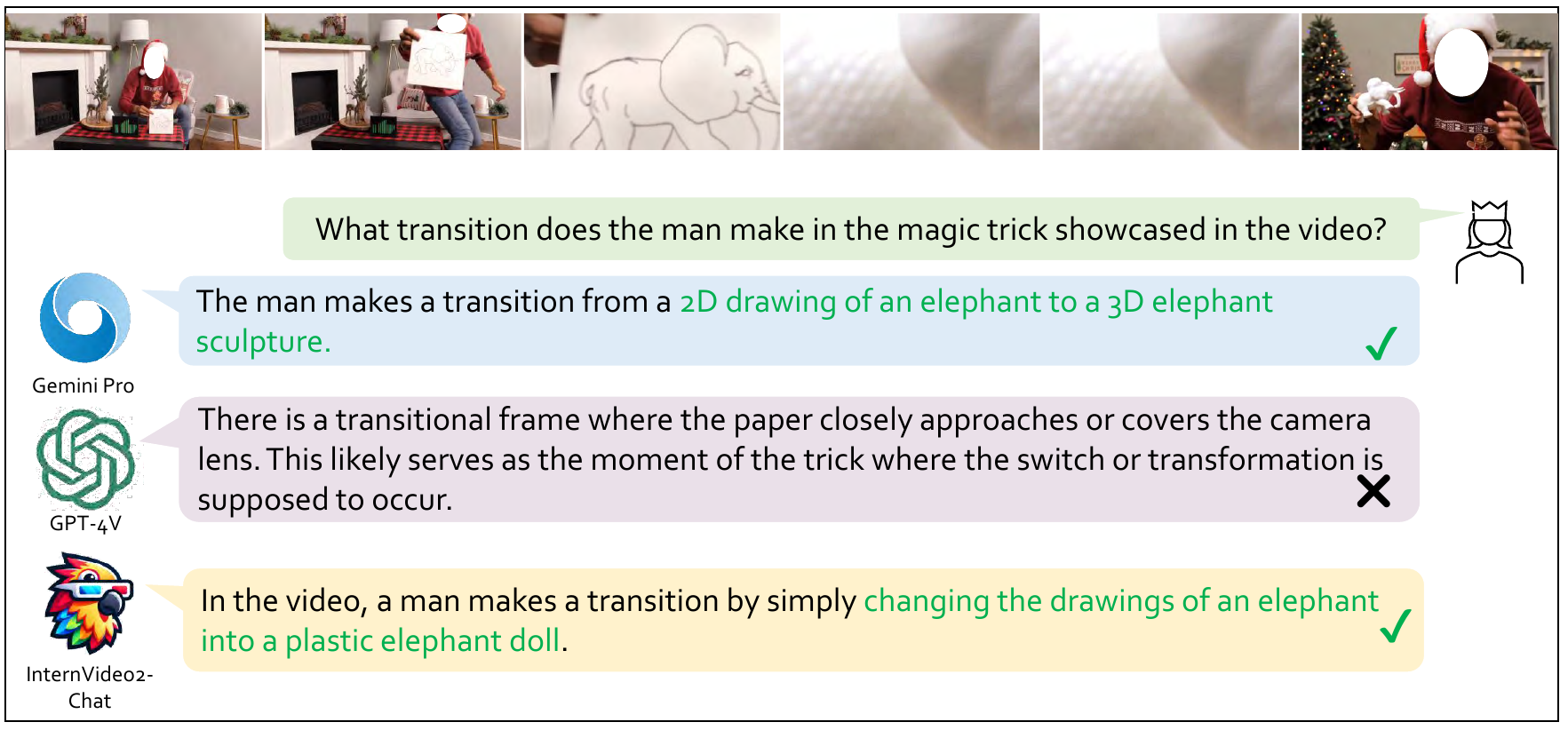}
    \caption{\textbf{Unexpected Action Recognition tasks.} The model needs to recognize the magical parts of the video. Both Gemini Pro and InternVideo2-Chat can capture part of the transition in the video and infer the shooting technique of the video. GPT-4V recognizes the transition but fails to successfully explain the process of the transition.}
    \label{fig:chat_reason}
\end{figure*}

\begin{figure*}[!t]
    \centering
    \includegraphics[width=0.9\textwidth]{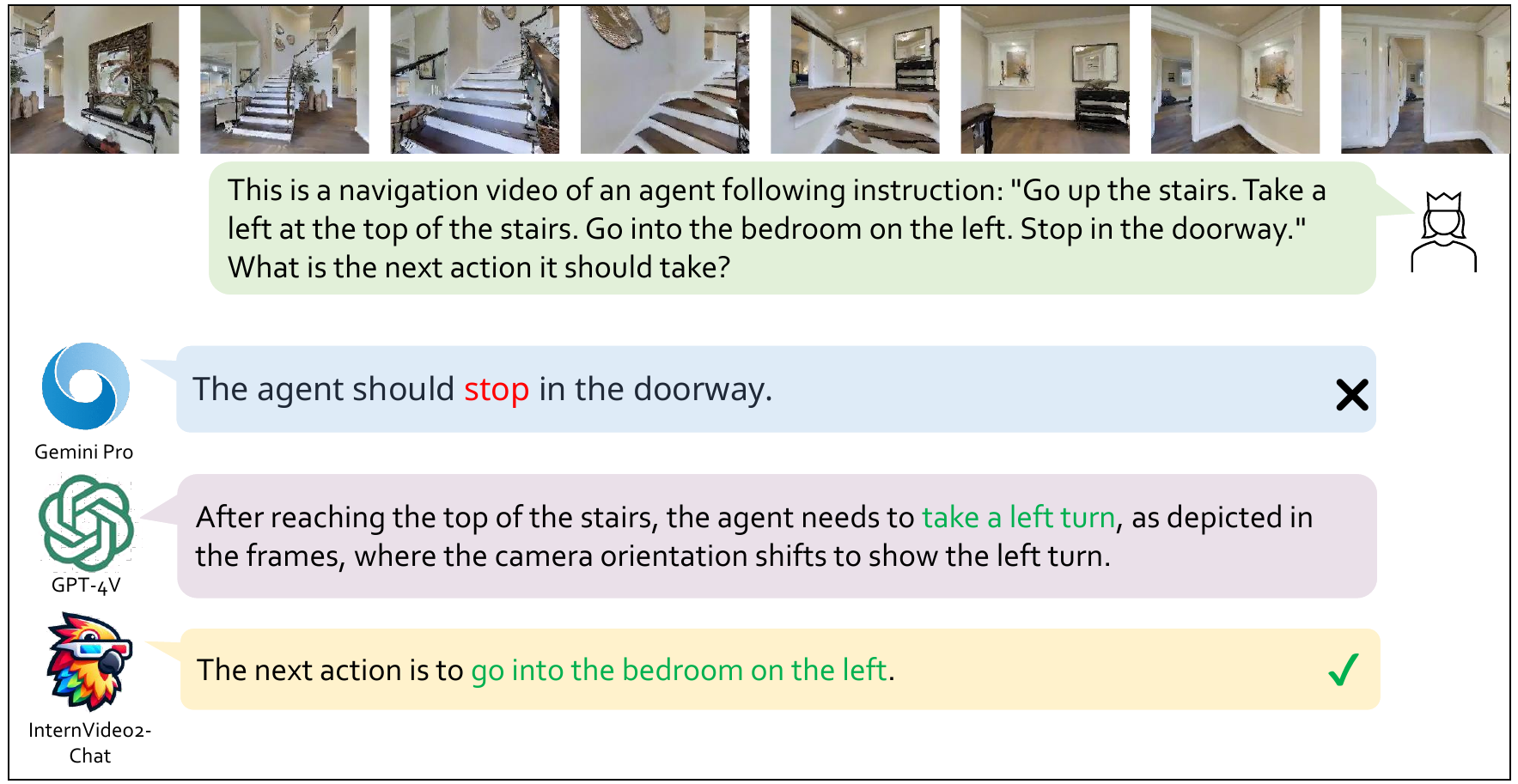}
    \caption{\textbf{Visual Language Navigation Tasks.} GPT-4V and InternVideo2-Chat are able to understand the instruction and make decisions about next steps based on the content of the video, while Gemini Pro is subject to hallucination.}
    \label{fig:chat_vln}
\end{figure*}

\subsection{Video-centric Dialogue and its Applications} \label{sec:chat}

Tab. \ref{tab:wct-vcbench} shows results on MVBench \cite{li2023mvbench}, Egoschema \cite{egoschema}, and Perception Test \cite{perceptiontest}, by equipping VideoChat2~\cite{li2023mvbench} with our InternVideo2 encoder, compared with other MLLMs. Note VideoChat2-HD-F16 with {\modelname} outperforms other systems with a clear margin in Perception Test and MVBench (\textbf{F16} denotes the model is trained and tested with 16-frame inputs), except in Egoschema compared with proprietary commercial models (GPT4 \& Gemini). Our method demonstrates superior short-term fine-grained video understanding compared to both open-source and commercial methods. Egoschema requires longer context harnessing, and we plan to further explore leveraging LLM for long multimodal sequence modeling. Considering these benchmarks not only involve perception but also reasoning, these results suggest {\modelname} does embed knowledge to partially model the world.
It also validates the importance of learning a transferrable video representation for current video-related MLLM. We also give several qualitative evaluations with popular GPT-4V and GeminiPro,
including dialogues on action sequence (Fig.~\ref{fig:chat_action}), confused action (Fig.~\ref{fig:chat_fine}), temporal order understanding (Fig.~\ref{fig:chat_order}), temporal event counting (Fig.~\ref{fig:chat_count}, unexpected action reasoning (Fig.~\ref{fig:chat_reason}), and vision-language navigation (Fig.~\ref{fig:chat_vln}).

\subsection{Ablation Studies} \label{sec:ab}

\subsubsection{Scaling Video Encoder} Tab. \ref{tab:model_scale} gives {\modelname}'s average performance on action recognition and video retrieval. It shows that scaling video encoder from 1B to 6B still leads to notable  improvements in generalization of action recognition and video-retrieval by 1.4\% and 1.9\% (in zero-shot), respectively. Meanwhile, the increase in finetuned action recognition result is relatively marginal (0.4\%) with the growth of model scale.

\subsubsection{Training Data and used Teachers in Stage 1} \label{sec:ab_kmash}
In Tab. \ref{tab:abaltion_stage1}, we examine the impact of distillation teachers and used dataset sizes on model performance.
\textit{(a) Data Scale:} 
Note that pretraining data scale for MAE should grow with the increasing model scale, otherwise the downstream performance would be saturated, such as K710 (0.66M videos) for ViT-L, K-Mash$_{1.1M}$ for ViT-1B, and K-Mash$_{2M}$ for ViT-6B.
\textit{(b) Teacher:} Tab. \ref{tab:abaltion_stage1} reveals that the synergy between a multimodal teacher (e.g., CLIP \cite{clip}) and a motion-aware teacher (e.g., MAE \cite{videomae}) markedly boosts performance, 
especially on SthSthV2. It highlights the importance of strategic teacher model selection in the distillation process.

\begin{table}[t]
\vspace{-2mm}
\caption{Ablation of Stage2. All models are tested with 8$\times$224$\times$224 input. 
}
\centering
\small
\resizebox{.75\linewidth}{!}{
\setlength{\tabcolsep}{3.0 mm}{
  \begin{tabular}{l|c}
  \toprule
Method & MSR-VTT \\
\midrule
Baseline (video \& text encoders w/ video-text learning) & 24.7 \\
Baseline + audio encoder w/ audio-text learning & 24.0\\
Baseline + speech encoder w/ video-speech-text learning & 24.9 \\
Baseline + audio encoder w/ video-audio-text learning & \textbf{27.8} \\
Baseline + audio \& speech encoders + video-audio-speech-text learning & 25.7 \\
\bottomrule
\end{tabular}
}
}
\label{tab:ab_stage2}
\end{table}

\begin{table}[t]
\vspace{-0mm}
\caption{Zero-shot t2v retrieval on MSR-VTT with different training captions. 
}
\centering
\small
\resizebox{.75\linewidth}{!}{
\setlength{\tabcolsep}{3.0 mm}{
\begin{tabular}{ccccc|c}
\toprule
SceneDet & AutoShot & Video Cap & Audio Cap & Speech Cap & MSR-VTT \\
\midrule
\Checkmark &  & \Checkmark &  &  & 24.7 \\
\Checkmark &  & \Checkmark & \Checkmark &  & 26.6 \\
 \Checkmark&  & \Checkmark & \Checkmark & \Checkmark & 27.1 \\
 & \Checkmark & \Checkmark & \Checkmark & \Checkmark & \textbf{34.8} \\
 \bottomrule
\end{tabular}
}
}
\label{tab:ab_stage2_data}
\end{table}

\subsubsection{Training Arch, Method, and Data in Stage 2} We ablate the necessity of introducing an audio encoder in Stage 2. We employ ViT-B and Bert-B for video and text encoders, respectively. The used text are simple video captions. The baseline is conducting video-text contrastive and matching as well as masked language generation loss for training with only video and text encoders. Other settings including adding audio or speech encoder or them both, and how to update newly added encoders i.e., whether train them with only text encoder or both video and text encoders. Tab. \ref{tab:ab_stage2} shows that only introducing audio encoder and learn it along with both video and text encoders can best improve video-text retrieval performance. The speech encoder harms such effectiveness more or less.

Further, we verify the impact of video temporal segmentation and the used captions as text inputs in Stage 2 in Tab. \ref{tab:ab_stage2_data}. 
We still use ViT-B and Bert-B for video-text training. 
Tab. \ref{tab:ab_stage2_data} finds the fused text from video-audio-speech works best for retrieval tasks compard with others, rising zero-shot t2v R\@1 of MSR-VTT from 24.7 to 27.1. Moreover, using AutoShot instead of SceneDet notably improves t2v retrieval (increasing by nearly 7 points). It validates the effectiveness of the introduced video-text dataset and its annotation systems.

\begin{table*}[!tp]
    \vspace{-3mm}
    \caption{
        Ablation on using qformer instruction in Stage3 training of Chat-Centric Model.
    }
    \centering
    \small
    \resizebox{0.86\linewidth}{!}{
    \setlength{\tabcolsep}{3.0 mm}{
        \begin{tabular}{l|l|l|c|c}
        \toprule

        \textbf{Model} & MVBench & NextQA & Egoschema-full & Egoschema-subset \\
        \midrule

        \textbf{VideoChat2} w qformer inst & 59.9 & \textbf{79} & 52.9 & 65.8 \\
        \textbf{VideoChat2} w/o qformer inst & \textbf{60.4} (+0.5) & 78.6 (-0.4)& \textbf{55.8} (+2.9) & \textbf{66.4} (+0.6)  \\
        
        \bottomrule
        \end{tabular}
        }
    }
    \vspace{-3mm}
    \label{tab:wct-qformerinst-ablate}
\end{table*}

\subsubsection{Training and Evaluation in Stage 3}
As shown in Tab. \ref{tab:wct-qformerinst-ablate}, incorporating questions (i.e., the `\texttt{q}' in `\texttt{QA}') into QFormer during Stage 3 training, which was found useful in \cite{instructblip}, actually harms the out-of-domain performance of the scaled-up VideoLLM. The NextQA training data is already included in the training corpus, and this is the only benchmark where the question-injected version performs better. Therefore, we believe that adding questions to QFormer during the instruction tuning stage of the scaled VideoChat model leads to some degree of overfitting.
\section{Conclusion and Discussion}
We have introduced a new family of video foundation models called {\modelname}, which achieves the state-of-the-art performance across various video and audio tasks. In {\modelname}, we combine masked video modeling, video-audio-text contrastive learning, and next token prediction into a unified framework. Additionally, we create a new video-text dataset that incorporates video-audio-speech fused captions as descriptions. The dataset contains temporally segmented clips with high semantic coherence.
These designs in {\modelname} contribute to enhancing video understanding in both perception and reasoning tasks. Notably, {\modelname} excels in video-related dialogue and long video understanding, demonstrating its effectiveness in capturing high-level semantics.

\paragraph{Limitations and Discussions.} Despite its achievements, {\modelname} does not introduce specific novel architectural design. Instead, it leverages the existing learning techniques for scaling video foundation models while focusing on improving data processing to enhance its spatiotemporal perception, semantic alignment, and basic knowledge embedding. Similarly to previous studies~\cite{umt,mplugowl}, InternVideo2 still grapples with limitations stemming from fixed input resolutions, sampling rates, and highly compressed tokens, which restrict its ability to express rich video information and capture fine-grained details.

The progressive learning scheme adopted by {\modelname} strikes a balance between model capabilities and training compute. While jointly learning the three optimization objectives simultaneously is computationally feasible, scalability becomes an issue when confronted with limited resources.

Although {\modelname} has demonstrated leading performance in some video understanding and reasoning benchmarks, it cannot guarantee an implicit world model that ensures consistency in visual reasoning. The inherent constraints imposed by fixed input representations, coupled with the complexity of visual reasoning tasks, present challenges in achieving a comprehensive and consistent understanding of the visual world.

\paragraph{Potential Biases.} We investigate the potential biases here. We focus on age, gender, and race distributions, as these are commonly recognized areas where bias can occur. 
We count keywords related to these categories in the used captions. Note that these synthetic captions may not fully reflect the truth of the corresponding videos, thereby creating a gap between our analysis and the actual reality. Here are the results of our analysis:
\begin{itemize}[leftmargin=*]
\item[$\bullet$] \textbf{Age}: The majority were about adults (86.99\%), followed by children (12.87\%) and barely any mentions of senior citizens (0.04\%).
\item[$\bullet$] \textbf{Gender}: 62.04\% pertained to men and 37.96\% pertained to women.
\item[$\bullet$] \textbf{Race}: 56.19\% are Asians, 23.04\% are Black people, 14.55\% are White people, 3.78\% are Middle Eastern people, and 2.43\% are Latin American people.
\end{itemize}

\section{Broader Impact}
It is important to acknowledge that, similar to other foundational models, {\modelname} has the potential to embed biases present in its training data and the associated models used during training, such as neural teachers \cite{videomae,chen2023internvl} and language models (LLMs) \cite{mistral,vicuna}. These biases may emerge due to a variety of factors, including the personal ideas, preferences, values, and perspectives of the data creators and the training corpus utilized.

The presence of biases in AI models can have societal implications and reinforce existing inequalities or prejudices. Biases within {\modelname} could manifest in the form of unfair or discriminatory outputs, potentially perpetuating social biases or stereotypes present in the training data. Consequently, it is crucial to carefully consider the potential impact of deploying {\modelname} in real-world applications and take proactive measures to mitigate biases and ensure fairness.

\bibliographystyle{plainnat}
\bibliography{references} 

\newpage
\appendix
\section{Model}

\begin{table*}[!th]
    \caption{Architecture of vision encoder (6B).}
    \vspace{-0.3cm}
    \centering
    \resizebox{0.9\linewidth}{!}{
        \begin{tabular}{c|c|c}
        \Xhline{1.0pt}
        \textbf{Stage} & \textbf{ViT-6B} & \textbf{Output Size} \\
        \Xhline{1.0pt}
        Video & sparse sampling & \violet{3}$\times$\darkGreen{8}$\times$\myblue{224}$\times$\myblue{224} \\
        \hline
        Patch & \darkGreen{1}$\times$\myblue{14}$\times$\myblue{14}, \violet{3200} & \multirow{2}{*}{\violet{3200}$\times$\darkGreen{8}$\times$\orange{256}} \\
        Embedding & stride \darkGreen{1}$\times$\myblue{14}$\times$\myblue{14} & ~ \\
        \hline
        Position & learnable, 3D sine-cosine initialization & \multirow{2}{*}{\violet{3200}$\times$\orange{2048}} \\
        Embedding & \violet{3200}$\times$\orange{2048} & ~ \\
        \hline
        Mask & semantic mask w/ \textit{mask ratio} $=$ $\rho$ & \violet{3200}$\times$\orange{2048$\cdot$(1-$\rho$)} \\
        \hline
        Encoder & $\left[\begin{array}{c}\text{MHSA(\violet{3200})}\\\text{MLP(\violet{12800})}\end{array}\right]$$\times$48 $+$ AttnPool(\violet{768}) & 
        \makecell{\violet{3200}$\times$\orange{2048$\cdot$(1-$\rho$)}
        \\\violet{768}$\times$\orange{1}} \\
        \hline
        Projection & 
        $\left[\begin{array}{c}\text{LN(\violet{3200})}\\\text{MLP(\violet{3200})}\end{array}\right]$$\times$$K_{CLIP}$,
        $\left[\begin{array}{c}\text{LN(\violet{3200})}\\\text{MLP(\violet{1408})}
        \end{array}\right]$$\times$$K_{MAE}$,
        $\left[\begin{array}{c}\text{LN(\violet{3200})}\\\text{MLP(\violet{768})}
        \end{array}\right]$$\times1$ & 
        \makecell{$K$$\times$\violet{3200}$\times$\orange{2048$\cdot$(1-$\rho$)}
        \\$K$$\times$\violet{1408}$\times$\orange{2048$\cdot$(1-$\rho$)}
        \\$K$$\times$\violet{768}$\times$\orange{1}} \\

        \bottomrule
        \end{tabular}
    }
    \label{tab:model_architecture}
    \vspace{-0.3cm}
\end{table*}

\paragraph{Video Encoder.}
In Tab. \ref{tab:model_architecture}, we take ViT-6B as an example and omit the class token for a simple presentation. ``MHSA'', ``MLP'', ``AttnPool'', and ``LN'' refer to spatiotemporal multi-head self-attention, multi-layer perception, attention pooling~\cite{coca} and root mean square layer normalization~\cite{zhang2019root}. $K_{CLIP}$ and $K_{MAE}$ means the layer number for unmasked token alignment with multimodal and motion-aware teachers. We mark the \violet{channel number}, \darkGreen{frame number}, \myblue{spatial size}, and \orange{token number} by different colors. The projection layers are dropped after stage 1 training.

\section{Video-centric Multimodal Data}
We prepare our training data according to the learning objectives of the three stages in our learning scheme. Specifically, it consists of video-only pretraining set for masked video token reconstruction, Video-Audio-Speech-Text one for multimodal alignment, and video instruction dataset for human-computer interaction alignment. They are detailed in the following.

\subsection{Video-only Data}
\begin{table}[tp]
    \centering
    \caption{Statistics of Stage1 data.
    All the videos are used without any labels.
    }
    \resizebox{0.67\linewidth}{!}{
        \begin{tabular}{l|c|c|c|c|c|c}
        \Xhline{1.0pt}
        \textbf{Dataset} & \textbf{K710} & \textbf{SthSthV2} & \textbf{HACS} & \textbf{ANet} & \textbf{MiT} & \textbf{Self-collected}\\
        \hline
        \textbf{K-Mash$_{1.1M}$} & 658K & 169K & 106K & 15K & 152K & 0 \\
        \textbf{K-Mash$_{2M}$} & 658K & 169K & 106K & 15K & 207K & 844K \\
        \Xhline{1.0pt}
        \end{tabular}
    }
    
    \label{tab:stage1_data}
    \vspace{-0.3cm}
\end{table}

To create the curated collection of videos, named \textbf{K-Mash}, we source videos from renowned action recognition datasets such as Kinetics-400 (K400) \cite{k400}, Something-Something (Sth) \cite{sth}, Moments in Time (MIT) \cite{mit}, ActivityNet \cite{activitynet}, and HACS \cite{hacs}. These datasets provide a wide range of video types, including both first-person and third-person perspectives, short and long durations, and featuring a rich variety of characters and settings.

For the enhanced version of the dataset, called K-Mash$_{2M}$, we push a step further and meticulously selected an additional 844,000 videos from YouTube to further enhance the diversity of the dataset. It's important to note that all videos in this dataset are utilized for training without any labels, allowing the model to learn from the unlabeled data in an unsupervised manner. This approach helps to broaden the model's understanding of different visual concepts and improves its performance on various video-related tasks. 

\subsection{Videos with Audio-Video-Speech Modalities} \label{supp:avs_anno}
In addtion to publicly available video-text datasets (e.g. InternVid~\citep{wang2023internvid} and WebVid \citep{webvid}), we introduce a new video dataset that incorporates both audio-visual-speech information and their corresponding textual descriptions. This dataset is included in the training process of {\modelname}. To create this multimodal dataset, named {\dataname}, we leverage several video sources and provide detailed annotations. {\dataname} includes videos with synchronized audio, visual, and speech information, along with their corresponding textual descriptions. This multimodal dataset enables the training of {\modelname} to better understand and capture the connections between different modalities, enhancing its performance in various video-related tasks that require audio, visual, and speech understanding.

\noindent\textbf{Collection.}
In the {\dataname} dataset, approximately half of the videos are sourced from YouTube, while the remaining videos are gathered from anonymous sources. This is to improve the diversity of dataset, as relying solely on YouTube may result in limited depth of the dataset.
Furthermore, to study the impact of video culture backgrounds on the learned models, a small portion of the dataset consists of Chinese data. These videos were collected with proper permissions for academic usage, ensuring compliance with legal and ethical considerations.

By incorporating videos from various sources and including a subset of Chinese data, {\dataname} provides a more diverse and representative dataset for training {\modelname}. This approach allows the model to learn from a wide range of video content, encompassing different cultural backgrounds and further enhancing its ability to understand and process videos from various sources.

\noindent\textbf{Trimming.}
In our approach, instead of relying on the widely-used SceneDet filter of FFMPEG, we employ a temporal boundary detection model called AutoShot \cite{zhu2023autoshot} to segment videos into clips. AutoShot is capable of predicting clip boundaries based on temporal semantic variations, as opposed to pixel differences. This leads to the generation of semantically complete cuts without mixing extra frames that may contain inconsistent context. By using AutoShot, we aim to reduce captioning errors by producing fewer clips with obvious transitions, resulting in a more coherent input for video captioning models. In the inference of AutoShot, we use a threshold of 0.5 to determine the shot boundaries for AutoShot's estimations. 

For the video dataset, we first preserve clips longer than 2 seconds. For video clips longer than 30 seconds, as the segments within the clip are from the same shot, we randomly choose a 30-second segment. During this process, we also abandon clips with still or extreme dynamics, such as browsing a photo gallery. 

\begin{table*}[h]\centering
\begin{minipage}{0.99\columnwidth}\vspace{0mm}    
     \caption{\textbf{Fusion prompt.}The above prompt is to generate 2 multi-modal captions, the following prompt is to generate 3 multi-modal captions.}
    \label{fusion_prompt}
    \vspace{-2mm}
    \centering
    \begin{tcolorbox} 
        \centering
        \hspace{-6mm}
        \begin{tabular}{p{0.99\columnwidth}}
        \hspace{1mm}
        \begin{minipage}{0.99\columnwidth}
        You are a text analysis expert. About one video, here is 1 vision caption: \violet{\texttt{vid\_cap}}, 1 audio caption:\violet{\texttt{aud\_cap}}. You need to understand and encode them into 1 sentence. Do not simply concatenate them together. The weights of video/audio are equaled. Considering dropping audio caption if it is incomprehensible. The output must be a complete and natural sentence. The sentence is:
        $...$ \\ \rule[0.25\baselineskip]{\textwidth}{1pt}
        You are a text analysis expert. About one video, here is 1 vision caption: \violet{\texttt{vid\_cap}}, 1 audio caption:\violet{\texttt{aud\_cap}}, and one speech subtitle: \violet{\texttt{asr\_cap}},. You need to understand and encode them into 1 compelete sentence. The weights of video/audio/speech are equaled. Considering dropping audio caption or speech subtitle if it is incomprehensible. The output must be a complete and natural sentence, do not simply concatenate them together. The complete sentence is:
        \end{minipage}
        \end{tabular}
    \end{tcolorbox}
    \vspace{-2mm}
\end{minipage}
\end{table*}

\begin{figure*}[t]
    \centering
    \includegraphics[width=1.\textwidth]{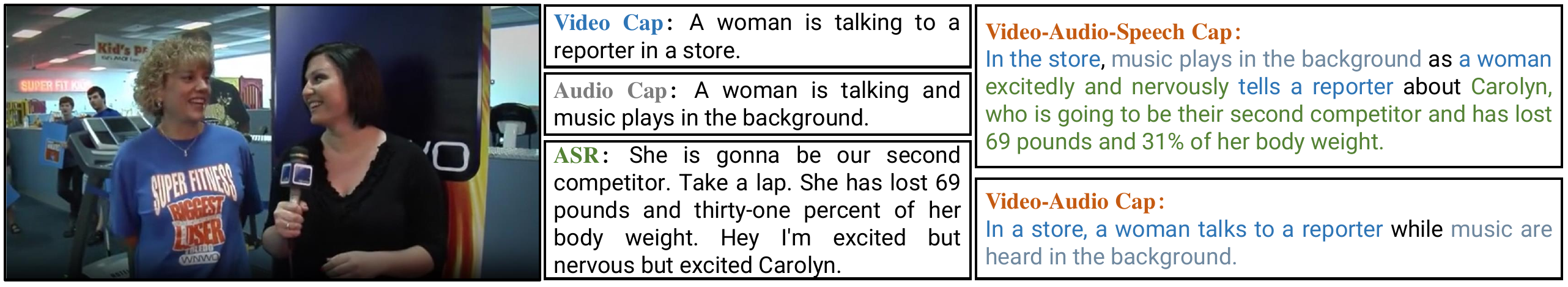}
    \includegraphics[width=1.\textwidth]{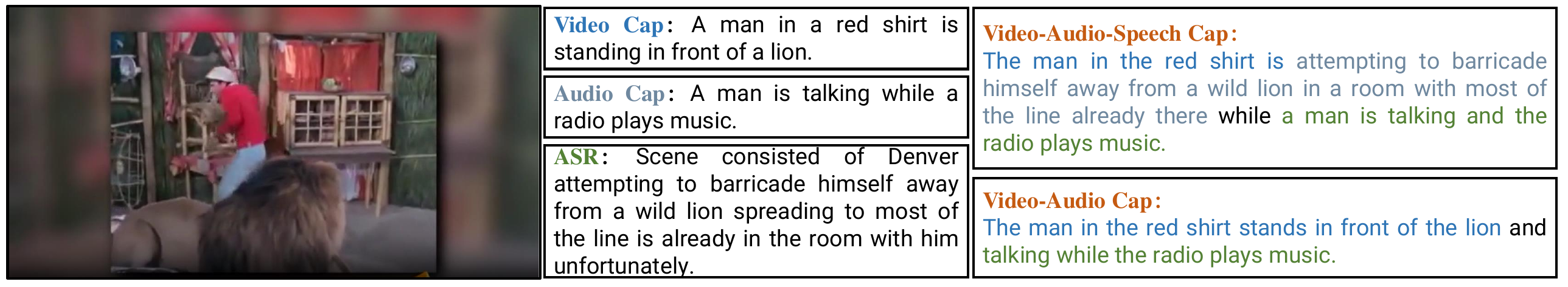}
    \includegraphics[width=1.\textwidth]{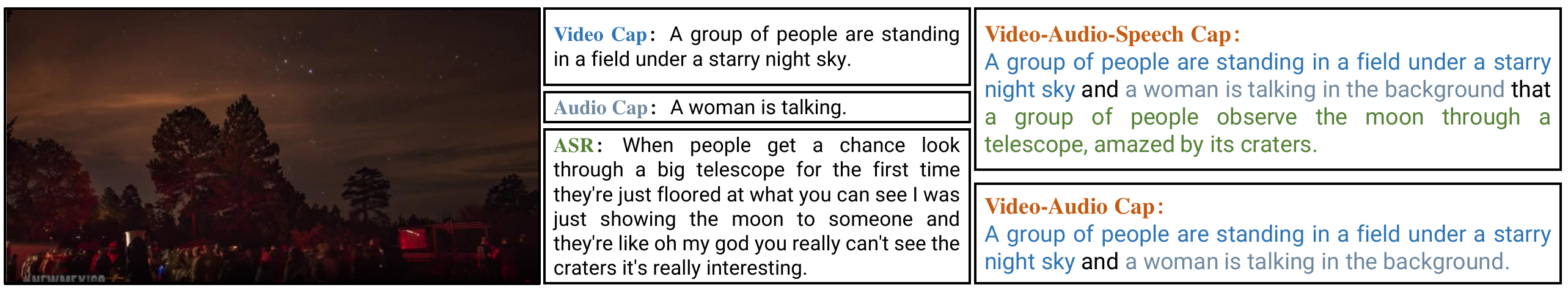}
    \includegraphics[width=1.\textwidth]{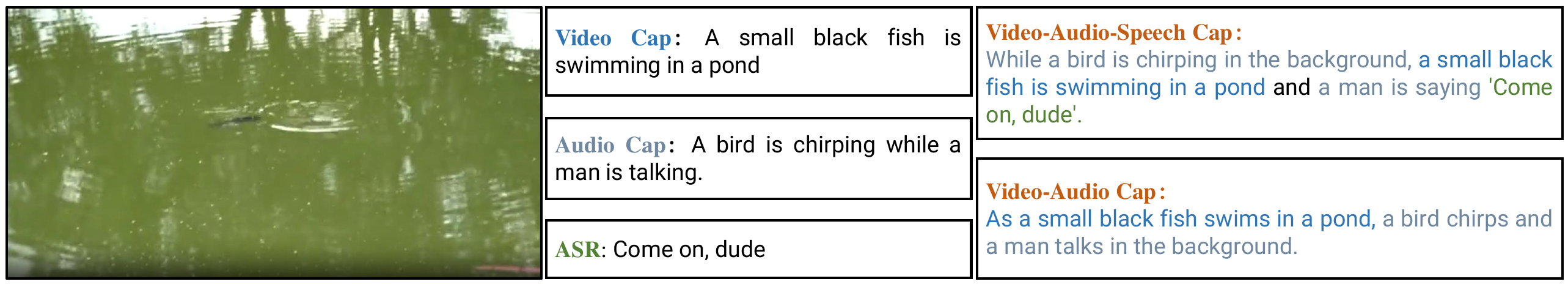}
    \caption{Annotation examples using our captioning approach.}
    \label{fig:anno_example}
\end{figure*}

\noindent\textbf{Annotation.} We automatically caption visual, audio, and speech of {\dataname}. Then we correct them and fuse them for cross-modal captions for training using LLM. We list several annotation examples of our method in Fig. \ref{fig:anno_example}.
\begin{itemize}[leftmargin=*]
    \item[$\bullet$] \textbf{Vision Captioner.} We employ the video captioning pipeline in InternVid to annotate our data. Rather than using VideoLLM \citep{li2023videochat,videochatgpt} to describe videos, we choose this validated method due to its better downstream results. 
    \item[$\bullet$] \textbf{Audio Captioner.} We craft a audio captioner upon VideoChat~\citep{li2023videochat}, as we find no reliable ones. It extracts audio features from inputs by Beats~\citep{beats}. We learn it by only tuning its Qformer (the interface between audio encoder and LLM) using a combination of the large-scale audio-text corpus WavCaps~\citep{mei2023wavcaps} dataset.
    \item[$\bullet$] \textbf{Speech Captioner.} We utilize the audio transcription model Whisper~\citep{whisper} to obtain speech from videos.Concretely, we use the WhisperV2-large model since its concurrent state-of-the-art performance. During the data collection process, a portion of the data is directly adopted from YT-Temporal-180M, which already has well-aligned timestamps and adjusted speech recognition content. The remaining data is first passed through a pre-trained language identification model Fasttext-lid~\citep{joulin2016bag} to determine the language category, and then transcribed the non-English text into English using the pretrained Seamless M4T~\citep{communication2023seamlessm4t} model. For the text with language identification confidence less than 0.95, we use Vicuna-1.5 as a translation alternative.
    \item[$\bullet$] \textbf{Caption Triming \& Fusion with LLM.} After obtaining the captions of audio, video, and speech from the given video, we utilize an LLM (Vicuna-1.5~\citep{vicuna}) to integrate the uni-modal captions into the multimodal ones. To fulfill the request for multiple contrastive objectives, we combine the audio caption with the video caption as the audio-visual caption, as well as integrate the audio, video, and speech captions as the audio-visual-subtitle captions. In this way, we acquire 5 types of captions (3 uni-modal captions (A, V, S) and 2 multi-modal captions (AV, AVS)) for each video automatically. Specifically, we have carefully designed prompt templates (Fig.~\ref{fusion_prompt}) and employed vLLM~\citep{kwon2023efficient} for inference acceleration, effectively get the visual caption, audio caption, subtitle, audio-visual caption and audio-visual-speech caption while maintaining a natural human-like subtitle style.
\end{itemize}

\noindent\textbf{Filtering \& Sampling.} After obtaining the captions, we calculate the CLIP similarity between the video segments and captions. We select the top 60 million data as the video segment data for \dataname{}. For LAION-2B, we only select samples with CLIP similarity in the top 158 million for training.

\section{Experiments}

\subsection{Ablations}

\subsubsection{How {\modelname} Works in Feature-based Tasks.} \label{supp:tal} We study which part of {\modelname}$_{s1}$'s predictions are suitable for feature-based tasks, i.e. temporal action localition. We adhere the same train and test protocols as in the main paper.

In Tab. \ref{supp:tal}, the most effective feature tends to be located within the last few layers of the video encoder. This observation aligns with the similarity between feature-based temporal tasks and linear probing classification, which is reasonable. We undertake comprehensive experiments to investigate the impact of features from various layers, as detailed in Table~\ref{tab:abla-tal-feature}. The results reveal the best features appear between the last 5-th layer and 7-th layer.

\begin{table}[h]
\centering
\caption{Effect of feature extracted from the last 7 layers.}
    \resizebox{1.0\linewidth}{!}{
        \begin{tabular}{c|cc|cc|cc|cc}
                \toprule
                 Layer Index & \multicolumn{2}{c}{THUMOS-14} & \multicolumn{2}{c}{ActivityNet} & \multicolumn{2}{c}{HACS Segment} & \multicolumn{2}{c}{FineAction} \\
                   & 1B@mAP& 6B@mAP& 1B@mAP& 6B@mAP& 1B@mAP& 6B@mAP & 1B@mAP& 6B@mAP  \\
                 \midrule
                 -1  & 67.9& 70.3& 39.0& 40.7 & 39.5& 42.1& 25.4 & 25.3  \\
                 -2  & 68.4 & 71.0 & 39.3 & 40.5 & 41.0& 42.7& 26.2 & 26.4  \\
                 -3 & 69.0 & 71.3 & 39.7 & 40.5 & 41.2& 42.7& 27.0 & 26.6  \\
                 -4 & 69.3 & 71.4 & 39.6 & 41.1 & 41.3& 43.1& 27.1 & 27.1  \\
                 -5 & \textbf{69.9} & 71.8 & 39.7  & 40.9 & \textbf{41.4} & 43.1 & \textbf{27.2} & \textbf{27.7}  \\
                 -6 & 69.6 & \textbf{72.0} & 39.6 & \textbf{41.2} & 40.7& \textbf{43.3} & 27.0& 27.7  \\
                 -7 & 69.5 & 71.9  & \textbf{40.0} & 41.1 & 40.6 & 42.8 & 26.9 & 27.7  \\
                \bottomrule
        \end{tabular}
        \label{tab:abla-tal-feature}
    }
\end{table}

\subsection{Video Retrieval}
We detail R@1, R@5, and R@10 of zero-shot video retrieval from {\modelname} in Tab. \ref{tab:msrvtt}
-\ref{tab:msvd} for reference.

\begin{table}[t]
    \caption{Video retrieval results on MSR-VTT, DiDeMo, LSMDC, ActivityNet, VATEX, and MSVD. We report R@1, R@5, and R@10. \#F denotes input frame number in eval.}
    \vspace{-0.2cm}
    \begin{subtable}[t]{.48\linewidth}
		\centering
		\caption{MSR-VTT}
    	\setlength{\tabcolsep}{2pt} %
		\renewcommand*{\arraystretch}{1.10}  %
		\vspace{-0.3\baselineskip}
		\resizebox{1.0\textwidth}{!}{
			\begin{tabular}{lccccccc}
            \toprule
             \multirow{2}{*}{Method}& \multirow{2}{*}{\#F} &\multicolumn{3}{c}{Text-to-Video} & \multicolumn{3}{c}{Video-to-Text} \\
             &  & R@1 & R@5 & R@10 & R@1 & R@5 & R@10\\
            \midrule
             \modelname$_{s2}$-1B & 4 & 51.9 & 74.6 & 81.7 & 49.6 & 73.6 & 81.2 \\
             \modelname$_{s2}$-1B & 8 & 51.9 & 75.3 & 82.5 & 50.9 & 73.4 & 81.8 \\      
             \modelname$_{s2}$-6B & 4 & 54.5 & 77.5 & 83.7 & 52.3 & 75.3 & 83.5 \\
             \modelname$_{s2}$-6B & 8 & 55.9 & 78.3 & 85.1 & 53.7 & 77.5 & 84.1 \\
            \bottomrule
          \end{tabular}
		}
		\label{tab:msrvtt}
  		\centering
  		\caption{LSMDC}
  		\vspace{-0.2\baselineskip}
  		\resizebox{1.0\textwidth}{!}{
  			\begin{tabular}{cccccccc}
            \toprule
             \multirow{2}{*}{Method}& \multirow{2}{*}{\#F} &\multicolumn{3}{c}{Text-to-Video} & \multicolumn{3}{c}{Video-to-Text} \\
             & & R@1 & R@5 & R@10 & R@1 & R@5 & R@10\\
            \midrule
            \modelname$_{s2}$-1B & 4 & 31.5 & 51.3 & 59.5 & 27.1 & 44.8 & 51.8 \\
             \modelname$_{s2}$-1B & 8 & 32.0 & 52.4 & 59.4 & 27.3 & 44.2 & 51.6 \\    
             \modelname$_{s2}$-6B & 4 & 34.8 & 54.0 & 61.6 & 30.1 & 48.0 & 55.0 \\
             \modelname$_{s2}$-6B & 8 & 33.8 & 55.9 & 62.2 & 30.1 & 47.7 & 54.8 \\
            \bottomrule
          \end{tabular}
  			\label{tab:lsmdc}
  		}
  		\centering
  		\caption{VATEX}
  		\vspace{-0.2\baselineskip}
  		\resizebox{1.0\textwidth}{!}{
  			\begin{tabular}{cccccccc}
            \toprule
             \multirow{2}{*}{Method}& \multirow{2}{*}{\#F} &\multicolumn{3}{c}{Text-to-Video} & \multicolumn{3}{c}{Video-to-Text} \\
             & & R@1 & R@5 & R@10 & R@1 & R@5 & R@10\\
            \midrule
             \modelname$_{s2}$-1B & 4 & 70.7 & 93.7 & 96.9 & 85.9 & 97.6 & 99.2\\
             \modelname$_{s2}$-1B & 8 & 70.4 & 93.4 & 96.9 & 85.4 & 97.6 & 99.1\\    
             \modelname$_{s2}$-6B & 4 & 71.1 & 93.8 & 97.0 & 85.2 & 97.7 & 99.4 \\
             \modelname$_{s2}$-6B & 8 & 71.5 & 94.0 & 97.1 & 85.3 & 97.9 & 99.3 \\
            \bottomrule
          \end{tabular}
  			\label{tab:vatex}
  		}
	\end{subtable}
    \begin{subtable}[t]{.48\linewidth}
		\centering
		\caption{DiDeMo}
    	\setlength{\tabcolsep}{2pt} %
		\renewcommand*{\arraystretch}{1.10}  %
		\vspace{-0.3\baselineskip}
		\resizebox{1.0\textwidth}{!}{
			\begin{tabular}{lccccccc}
            \toprule
             \multirow{2}{*}{Method}& \multirow{2}{*}{\#F} &\multicolumn{3}{c}{Text-to-Video} & \multicolumn{3}{c}{Video-to-Text} \\
             &  & R@1 & R@5 & R@10 & R@1 & R@5 & R@10\\
            \midrule
             \modelname$_{s2}$-1B & 4 & 56.7 & 78.7 & 83.9 & 54.4 & 74.4 & 80.6 \\
             \modelname$_{s2}$-1B & 8 & 57.0 & 80.0 & 85.1 & 54.3 & 77.2 & 83.5 \\    
             \modelname$_{s2}$-6B & 4 & 56.2 & 77.6 & 83.6 & 53.2 & 76.8 & 82.7 \\
             \modelname$_{s2}$-6B & 8 & 57.9 & 80.0 & 84.6 & 57.1 & 79.9 & 85.0 \\
            \bottomrule
          \end{tabular}
		}
		\label{tab:didemo}
  		\centering
  		\caption{ActivityNet}
  		\vspace{-0.2\baselineskip}
  		\resizebox{1.0\textwidth}{!}{
  			\begin{tabular}{cccccccc}
            \toprule
             \multirow{2}{*}{Method}& \multirow{2}{*}{\#F} &\multicolumn{3}{c}{Text-to-Video} & \multicolumn{3}{c}{Video-to-Text} \\
             & & R@1 & R@5 & R@10 & R@1 & R@5 & R@10\\
            \midrule
             \modelname$_{s2}$-1B & 4 & 56.9 & 81.7 & 89.8 & 53.6 & 80.0 & 88.5\\
             \modelname$_{s2}$-1B & 8 & 60.4 & 83.9 & 90.8 & 54.8 & 81.5 & 89.5 \\    
             \modelname$_{s2}$-6B & 4 & 59.4 & 83.2 & 90.3 & 53.7 & 80.5 & 88.9 \\
             \modelname$_{s2}$-6B & 8 & 63.2 & 85.6 & 92.5 & 56.5 & 82.8 & 90.3 \\
            \bottomrule
          \end{tabular}
  			\label{tab:anet}
  		}
  		\centering
  		\caption{MSVD}
  		\vspace{-0.2\baselineskip}
  		\resizebox{1.0\textwidth}{!}{
  			\begin{tabular}{cccccccc}
            \toprule
             \multirow{2}{*}{Method}& \multirow{2}{*}{\#F} &\multicolumn{3}{c}{Text-to-Video} & \multicolumn{3}{c}{Video-to-Text} \\
             & & R@1 & R@5 & R@10 & R@1 & R@5 & R@10\\
            \midrule
             \modelname$_{s2}$-1B & 4 & 58.9 & 83.0 & 88.7 & 83.6 & 94.8 & 97.0\\
             \modelname$_{s2}$-1B & 8 & 58.1 & 83.0 & 88.4 & 83.3 & 94.3 & 96.9\\    
             \modelname$_{s2}$-6B & 4 & 59.8 & 84.2 & 89.7 & 82.5 & 94.6 & 97.2\\
             \modelname$_{s2}$-6B & 8 & 59.3 & 84.4 & 89.6 & 83.1 & 94.2 & 97.0\\
            \bottomrule
          \end{tabular}
  			\label{tab:msvd}
  		}
	\end{subtable}
\end{table}

\begin{table}[!ht]
\vspace{-0mm}
\caption{The top-1 accuracy of zero-shot video QA (multi-choice) on MSR-VTT and LSMDC. Finetuned results are marked in gray.}
\centering
\small
\resizebox{0.50\textwidth}{!}{
\setlength{\tabcolsep}{1.5 mm}{
\begin{tabular}{lcc}
\toprule
Method & MSR-VTT & LSMDC \\
\midrule
VIOLET \cite{violet} & \gray{91.9} & \gray{82.8} \\
InternVideo \cite{wang2022internvideo} & 93.4 & \textbf{77.3}  \\
\modelname$_{s2}$-6B & \textbf{94.4} & 76.9 \\
\bottomrule
\end{tabular}}}
\label{tab:mc}
\end{table}

\subsection{Multi-Choice Video Question Answering} \label{sec:qa}
We evaluate zero-shot multi-choice (MC) video QA using {\modelname} in stage 2 on MSR-VTT and VATEX. 
Tab. \ref{tab:mc} shows {\modelname} consistently improves MC accuracy compared with previous SOTAs except on LSMDC, where it gets a comparable result with InternVideo.

\begin{table}[!t]
\caption{Performance of multimodal LLMs on different question types and scenes of MoVQA.
}
\centering
\scalebox{.65}{
\begin{tabular}{c|c|cccccc|ccc|c}
\toprule 
  Method & Backbone &Synopsis & Temporal & Spatial & Causal & Hypothetical  & Knowledge   & Single-Scene & Multi-Scene & Full-Scene & Overall \\ 
\midrule
  Mplug-Owl \cite{mplugowl}& CLIP ViT-L &25.1 &  19.9 & 25.3 & 21.9& 23.5  &27.5  & 25.2 & 23.5 & 22.1 & 24.8\\
  Otter \cite{li2023otter} & CLIP ViT-L & 22.6 & 20.7 &  19.6  &  26.1 & 24.2  & 21.8 & 23.1 & 22.1 & 21.3 & 22.6\\
  VideoChatGPT \cite{videochatgpt} & CLIP ViT-L &  23.8 & 20.2  &  22.1   &  22.1 &  21.4  &  24.1 & 23.4 & 22.7 & 22.3 & 22.9\\
  VideoChat \cite{li2023videochat} & Eva-g  &  33.6 & 24.3 & 34.5 & 36.6 & 35.5  & 32.0 &35.3 & 32.9 & 33.3 & 34.7\\ 
  VideoChat2 \cite{li2023mvbench} & {\modelname}$_{s3}$ & \bf42.6 & \bf27.8 & \bf39.9  & \bf44.3 & 4\bf42.5 & \bf41.2 & \bf40.9 & \bf39.3 & \bf38.6  & \bf40.1\\
\bottomrule
\end{tabular}
}
\label{tab:movqa-type-scene}
\end{table}

\subsection{Movie Understanding}

We evaluate {\modelname} on the MoVQA for movie understanding. It is a long-form movie question-answering dataset\cite{zhang2023movqa}. MoVQA assesses the diverse cognitive capabilities of multimodal systems by considering both video length and clue length, relying on multi-level temporal lengths (single-scene, multi-scene and full-scene). 
There are six types of QAs, including information synopsis, temporal perception, spatial perception, causal reasoning, hypothetical reasoning and external knowledge.
We evaluate {\modelname}-6B in the form of open-ended QAs, and detailed results are shown on Table~\ref{tab:movqa-type-scene}.

\end{document}